\documentclass[runningheads]{llncs}
\usepackage{graphicx}

\usepackage{tikz}
\usepackage{comment}
\usepackage{amsmath,amssymb} %
\usepackage{color}

\usepackage[accsupp]{axessibility}  %

\usepackage{caption}
\usepackage{subcaption}

\newcommand{\pic}[3]{
	\begin{figure}[tb]
		\centering
		\includegraphics[width=#3\linewidth]{#1}
		\caption{#2}
		\label{fig:#1}
	\end{figure}
}

\newcommand{\picTwo}[4]{
	\begin{figure*}[tb]
		\centering
		\begin{subfigure}[c]{0.46\textwidth}
			\centering
			\includegraphics[width=0.9\textwidth]{#1-0}
			\subcaption{#2}
			\label{fig:#1-0}
		\end{subfigure}
		\begin{subfigure}[c]{0.52\textwidth}
			\centering
			\includegraphics[width=0.9\textwidth]{#1-1}
			\subcaption{#3}
			\label{fig:#1-1}
		\end{subfigure}
		\caption{#4}
		\label{fig:#1}
	\end{figure*}
}

\newcommand{\picTwoExtended}[6]{
	\begin{figure*}[htb]
		\begin{subfigure}[c]{0.49\textwidth}
			\includegraphics[width=\textwidth]{#1-0}
				\subcaption{#2}
		\end{subfigure}
		\begin{subfigure}[c]{0.49\textwidth}
			\includegraphics[width=\textwidth]{#1-1}
				\subcaption{#3}
		\end{subfigure}
		\begin{subfigure}[c]{0.49\textwidth}
			\includegraphics[width=\textwidth]{#1-2}
				\subcaption{#4}
		\end{subfigure}
		\begin{subfigure}[c]{0.49\textwidth}
			\includegraphics[width=\textwidth]{#1-3}
				\subcaption{#5}
		\end{subfigure}
		\caption{#6}
		\label{fig:#1}
	\end{figure*}
}

\newcommand{\picFour}[6]{
	\begin{figure*}[tb]
		\begin{subfigure}[c]{0.24\textwidth}
			\includegraphics[width=\textwidth]{#1-0}
				\subcaption{#2}
				\label{fig:#1-0}
		\end{subfigure}
		\begin{subfigure}[c]{0.24\textwidth}
			\includegraphics[width=\textwidth]{#1-1}
				\subcaption{#3}
				\label{fig:#1-1}
		\end{subfigure}
		\begin{subfigure}[c]{0.24\textwidth}
			\includegraphics[width=\textwidth]{#1-2}
				\subcaption{#4}
				\label{fig:#1-2}
		\end{subfigure}
		\begin{subfigure}[c]{0.24\textwidth}
			\includegraphics[width=\textwidth]{#1-3}
				\subcaption{#5}
				\label{fig:#1-3}
		\end{subfigure}
		\caption{#6}
		\label{fig:#1}
	\end{figure*}
}

\usepackage{multirow}
\usepackage{booktabs}
\usepackage{colortbl}
\definecolor{LightGray}{rgb}{0.95,0.95,0.95}

\newcommand{\tblDatasets}{
\begin{table*}[tb]

		\caption{
		Overview of the used datasets --
		\# is an abbreviation for number.
		The class imbalance is given as the percentage of the smallest and largest class with regard to the complete dataset.
		$\hat p_A$ is the expected prior ambiguity probability of the dataset. $n$ is the average of annotations per image.
	}
	\label{tbl:datasets}

	\resizebox{\textwidth}{!}{
		\begin{tabular}{l c c c c c c c c c }
			\toprule
			Name & \# classes & Input size $\left[ px \right]$ & \multicolumn{3}{c}{\# Images} & \multicolumn{2}{c}{Class Imbalance $\left[ \% \right]$ } & $\hat p_A$ $\left[ \% \right]$ & n\\

			\cmidrule(r){4-6} \cmidrule(r){7-8}

			& & & Train &  Val & Unlabeled & Smallest & Largest  &  \\

			\midrule

			Plankton \cite{foc} & 10 & 96x96 & 1964 & 2456 & 7860 & 4.16 & 30.37 & 44 & 24\\
			Turkey \cite{turkey_dataset} & 2 & 96x96 & 1299 & 1542 & 5199 & 9.66 & 90.33 & 22 & 3 \\
			Mice Bone \cite{schmarje2019} & 3 & 224x224 & 277 & 169 & 278 & 10.81 & 63.98 & 65 & 3 \\
			CIFAR-10H \cite{cifar10h} & 10 & 32x32 & 1600 & 2000 & 6400 & 9.88 & 10.16  & 32 & 51\\

		\end{tabular}
	}

\end{table*}
}

\newcommand{\tblOrthogonalvTwo}{
\begin{table*}[tb]

	\caption{
		Performance across different methods and datasets --
		The vanilla algorithm is highlighted in light grey.
		Better results in comparison to the vanilla algorithm are marked bold.
		The definition of the metrics are given in \autoref{subsec:metric}.
		CE stands for supervised Cross-Entropy training.
		All values are given in \%.
		Reasons for exclusion:  H - Hardware Restrictions
	}
	\label{tbl:orthogonal}

	\resizebox{\textwidth}{!}{
		\begin{tabular}{l c c c c c c c c c c c c c} %
			\toprule
			Methods & \multicolumn{3}{c}{Plankton}  & \multicolumn{3}{c}{Turkey}   &  \multicolumn{3}{c}{Mice Bone} &  \multicolumn{3}{c}{CIFAR-10H} & STL-10 \\ %

			\cmidrule(r){2-4} \cmidrule(r){5-7} \cmidrule(r){8-10}    \cmidrule(r){11-13} \cmidrule(r){14-14} %

			& F1 $\uparrow$ & \disx $\downarrow$  &  \diffx $\downarrow$  &  F1 $\uparrow$ & \disx $\downarrow$  &  \diffx $\downarrow$ & F1 $\uparrow$ & \disx $\downarrow$  &  \diffx $\downarrow$ & F1 $\uparrow$ & \disx $\downarrow$  &  \diffx $\downarrow$ & F1 $\uparrow$  \\

			\midrule

			\rowcolor{LightGray}

			CE   &
			86.71 & 30.45  & -56.26 & 83.84 & 42.98 & -40.86  & 69.55 & 54.75 & -14.80 & 67.71 & 55.80 & -11.91 & 80.48 \\

			CE + DC3  &
			78.24  &  \textbf{23.41} & -54.84 & \textbf{85.79} & \textbf{27.64} & \textbf{-58.14} & \textbf{93.88} & \textbf{36.58} & \textbf{57.30} &  \textbf{78.27} & \textbf{54.52} & \textbf{-23.75} & \textbf{88.45} \\
			 \rowcolor{LightGray}
			 Mean-Teacher  \cite{mean-teacher}   &
			 88..72 & 25.84  & -62.88 & 81.82 & 45.12 & -36.70  & 66.41 & 48.83 & -17.58 & 73.53 & 46.93 & -26.59 & 80.67  \\

			Mean-Teacher \cite{mean-teacher} + DC3  &
			 \textbf{91.30} & \textbf{24.84} & \textbf{-66.46} & \textbf{86.45} & \textbf{33.92} & \textbf{-52.53}  & \textbf{89.4} & \textbf{35.11} & \textbf{-54.73} & \textbf{85.13} & 52.44 & \textbf{-32.69} & \textbf{89.28}\\

			 \rowcolor{LightGray}

			 Pi-Model \cite{temporal-ensembling}  &
			 87.57 & 28.43  & -59.14 & 82.11 & 39.46 & -42.65  & 68.15 & 54.11 & -14.04 & 71.53 & 49.13 & -22.40 & 82.56\\

			Pi-Model \cite{temporal-ensembling} + DC3  &
			79.79 & \textbf{19.08}  & \textbf{-60.71} & \textbf{87.43} & \textbf{23.33} & \textbf{-64.10}  & \textbf{88.01} & \textbf{30.99} & \textbf{-57.02} & \textbf{83.05} & \textbf{43.40} & \textbf{-39.65} & \textbf{89.54} \\

			 \rowcolor{LightGray}
			  Pseudo-Label \cite{pseudolabel}   &
			87.62 & 27.42  & -60.20 & 82.37 & 44.88 & -37.49  & 66.60 & 57.03 & -9.57 & 69.70 & 53.30 & -16.40 & 82.48 \\

			 Pseudo-Label \cite{pseudolabel} + DC3    &
			 \textbf{89.31} & 31.76  & -57.55 &\textbf{ 83.44} & \textbf{35.04} & \textbf{-48.41}  & \textbf{86.58} & \textbf{37.52} & \textbf{-49.06} & \textbf{83.74} & \textbf{51.32} & \textbf{-32.42 } & \textbf{88.87}\\

			\rowcolor{LightGray}
			  FixMatch \cite{fixmatch}   &
			85.81  & 30.29 & -55.52  & 82.14 & 43.33 & -38.81 & H & H & H & 78.09 & 41.99 & -36.10 & 89.35\\

			 FixMatch \cite{fixmatch} + DC3    &
			 \textbf{87.20} & 31.28 & \textbf{-55.92} & \textbf{83.56} & \textbf{28.17} & \textbf{-55.39} & H & H & H & \textbf{83.09} & 49.49 & -33.60 & \textbf{91.45}\\

		\end{tabular}
	}

\end{table*}
}

\newcommand{\tblConsistency}{
\begin{table}[tb]
	\centering
				\captionof{table}{
		Consistency comparison of generated labels from proposals --
		The first column describes the annotator selection and the used proposals.
		The Cohen's kappa coefficient $\kappa$ measures the agreement of between the used repetitions and Time gives annotation time in minutes.
		Results which are within one percent or minute of the best result per dataset and annotator selection are marked bold.
		}
	\label{tbl:cons}

	\resizebox{\linewidth}{!}{
		\begin{tabular}{l c c  c c c c c c c }
			\toprule
			 & \multicolumn{2}{c}{Plankton} &  \multicolumn{2}{c}{Turkey} &  \multicolumn{2}{c}{Mice Bone} &  \multicolumn{2}{c}{CIFAR-10H}  \\

			\cmidrule(r){2-3} \cmidrule(r){4-5} \cmidrule(r){6-7} \cmidrule(r){8-9}
			 &  $\kappa$ [\%] $\uparrow$  & Time [min] $\downarrow$ &  $\kappa$ [\%] $\uparrow$  & Time [min] $\downarrow$ &  $\kappa$  [\%] $\uparrow$  & Time [min] $\downarrow$ &  $\kappa$ [\%] $\uparrow$  & Tim [min] $\downarrow$ \\
		\midrule

		 	\rowcolor{LightGray}
        A1 & 73.00 $\pm$ 1.51 & 51.09 $\pm$ 2.36 & 88.08 $\pm$ 3.43 & 14.56 $\pm$ 0.84 & 71,35 $\pm$ 2.56 & 13,94 $\pm$ 2.25 & 92.70 $\pm$ 1.69 & 40.58 $\pm$ 1.93 \\
        A1 + SSL & 85.00 $\pm$ 2.52 & 12.69 $\pm$ 3.37  & 85.63 $\pm$ 3.66 & \textbf{10.70 $\pm$ 0.44} & 72.00 $\pm$ 2.87 & \textbf{6.59 $\pm$ 1.65} & \textbf{94.85 $\pm$ 0.91} & \textbf{14.33 $\pm$ 1.48}\\
        A1 + DC3 & \textbf{90.29 $\pm$ 1.41} & \textbf{11.32 $\pm$ 1.43} & \textbf{91.95 $\pm$ 1.22} & \textbf{11.57 $\pm$ 0.64} & \textbf{81.36 $\pm$ 2.17} & \textbf{6.74 $\pm$ 1.05} & \textbf{94.70 $\pm$ 0.52} & \textbf{14.65 $\pm$ 0.60}\\

      	\rowcolor{LightGray}
        A2  &  85.25 $\pm$ 1.79  &  61.99 $\pm$ 10.98  & \textbf{81.54 $\pm$ 0.89} & 18.11 $\pm$ 4.30 & 68.63 $\pm$ 6.66 & 11.06 $\pm$ 3.60 & \textbf{98.81 $\pm$ 0.14} & 33.08 $\pm$ 5.36\\
        A2 + SSL & \textbf{94.88 $\pm$ 0.52} & \textbf{9.23 $\pm$ 0.76} & \textbf{81.10 $\pm$ 3.39} & \textbf{9.48 $\pm$ 0.83} & 59.63 $\pm$ 6.20 & 12.07 $\pm$ 4.77  & \textbf{98.00 $\pm$ 0.27} & \textbf{12.66 $\pm$ 0.69} \\
        A2 + DC3 & \textbf{94.04 $\pm$ 0.67} & 10.32 $\pm$ 0.07  & \textbf{81.83 $\pm$ 1.98} &  \textbf{ 9.91 $\pm$ 0.39}  & \textbf{72.19 $\pm$ 3.23} & \textbf{9.13 $\pm$ 2.98} & \textbf{98.29 $\pm$ 0.19} & 14.27 $\pm$ 0.69 \\

        	\rowcolor{LightGray}
        A3 &  84.74 $\pm$ 1.02  & 21.54 $\pm$ 1.54   & 78.27 $\pm$ 1.08  & 19.35 $\pm$ 1.16 & 56.27 $\pm$ 4.03 & 10.15 $\pm$ 2.12  & 93.22 $\pm$ 1.01 &  21.96 $\pm$ 1.10  \\
        A3 + SSL & \textbf{88.59 $\pm$ 0.84} & 9.02 $\pm$ 0.20 & 88.44 $\pm$ 1.74 & \textbf{13.24 $\pm$ 0.32} & \textbf{72.32 $\pm$ 0.61} & \textbf{8.02 $\pm$ 1.23} & 92.37 $\pm$ 1.78 & \textbf{ 9.79 $\pm$ 0.52 } \\
        A3 + DC3 & \textbf{88.57 $\pm$ 0.62}   & \textbf{7.76 $\pm$ 0.27} & \textbf{91.94 $\pm$ 1.04} & \textbf{14.05 $\pm$ 0.51} & \textbf{72.77 $\pm$ 2.74} & 9.56 $\pm$ 1.71  & \textbf{94.81 $\pm$ 0.96} & \textbf{9.50 $\pm$ 0.74} \\

		\end{tabular}
		}

\end{table}
}

\newcommand{\tblImpactFuzziness}{
\begin{table}[tb]
    \centering

    	\caption{
		Impact of ambiguous labels -- Macro F1-Score for different methods and across three different subsets on the validation data from the Plankton and CIFAR-10H dataset. Columns:
		A1 -- Labels are sampled from $\hat l$;  A -- Labels are the maximum class of $\hat l$;  C -- No ambiguous labels/images are used
	}

		\begin{tabular}{l c c c c c c }
			\toprule
			Methods & \multicolumn{3}{c}{Plankton} &  \multicolumn{3}{c}{CIFAR-10H}  \\

			\cmidrule(r){2-4} \cmidrule(r){5-7}

			&  A1 &  A  & C & A1 & A & C \\

			\midrule

				\rowcolor{LightGray}

			CE   &
			86.71 & 88.35 &  96.10 & 67.71 & 68.89 & 86.57 \\

			Mean-Teacher \cite{mean-teacher} &
			   88.72  & 88.94  &  96.00 & 73.56 & 75.06 & 86.96\\

			 \rowcolor{LightGray}

			 Pi-Model \cite{temporal-ensembling}  &
			  87.57  & 89.03 & 96.41 & 71.53 & 72.75 & 87.19\\

			  Pseudo-Label \cite{pseudolabel}   &
				  87.62  & 88.41 &  96.20 & 69.70 & 71.82 & 87.15 \\

		      FixMatch \cite{fixmatch}   &
				 80.29  & 90.24 & 98.86 & 76.15 & 79.15 & 90.37 \\

		\end{tabular}

	\label{tbl:fuzziness}
\end{table}
}

\newcommand{\tblAblationAverage}{
\begin{table*}[tb]

	\caption{
	    Ablation results averaged over different methods --
		The vanilla algorithms / baselines are highlighted in light grey.
    	Each lower row extends this baseline individually with CE$^{-1}$ \cite{foc}, Clustering \& Classification (CC) or both (DC3).
    	CC can be interpreted as DC3 without CE$^{-1}$.
    	The prior ambiguity estimate $p_A$ is given in brackets if applicable.
    	Results that improve over the baseline are marked in bold.
			The metrics are defined in \autoref{subsec:metric}.
			The column 'Ambiguous' gives the percentage of predicted \fuzzy data and the last column gives the number of non-degenerated runs over which we averaged
	}
	\label{tbl:ablation-average}

	\resizebox{\textwidth}{!}{
			\begin{tabular}{l c c c c c c c c c c}
				\toprule
			& \multicolumn{2}{c}{F1}  & \multicolumn{2}{c}{\disx}  & \multicolumn{2}{c}{\diffx} & \multicolumn{2}{c}{ Ambiguous } & \# Runs\\

\cmidrule(r){2-3} \cmidrule(r){4-5} \cmidrule(r){6-7}  \cmidrule(r){8-9}

				& best & mean $\pm$ std  &  best & mean $\pm$ std   &  best & mean $\pm$ std &  best  & mean $\pm$ std  & \\

			\midrule

			\textbf{CIFAR-10H}  & & & & & & \\

				\rowcolor{LightGray}

				Baseline  &
				0.7809 & 0.7153 $\pm$ 0.0359 & 0.4199 & 0.5027 $\pm$ 0.0469 & -0.3611 & -0.2126 $\pm$ 0.0827 & - & - & 15 \\
				 + CE$^-1$  &
				0.7383 & \textbf{0.7191 $\pm$ 0.0164} & 0.4692 & \textbf{0.4929 $\pm$ 0.0243} & -0.2691 & \textbf{-0.2262 $\pm$ 0.0404} &  - & - & 12 \\
				 + CC  ($p_A = 0.6$) &
				\textbf{0.8565} & \textbf{0.7471 $\pm$ 0.1246} & 0.8657 & 0.8768 $\pm$ 0.0129 & 0.0092 & 0.1297 $\pm$ 0.1374 & 0.6145 & 0.5923 $\pm$ 0.0322 & 12 \\
				 + DC3 ($p_A = 0.32$)   &
				0.6656 & 0.6970 $\pm$ 0.0469 & \textbf{0.2155} &\textbf{ 0.3684 $\pm$ 0.1227} & \textbf{-0.4501} & \textbf{-0.3286 $\pm$ 0.0836} & 0.2910 & 0.3115 $\pm$ 0.0140 & 12 \\
				 + DC3 ($p_A = 0.6$)  &
				\textbf{0.8305} & \textbf{0.7457 $\pm$ 0.1097} & 0.4340 &\textbf{ 0.4741 $\pm$ 0.0584} & \textbf{-0.3965} & \textbf{-0.2716 $\pm$ 0.0928} &0.6125 & 0.5860 $\pm$ 0.0290 & 15 \\
				\midrule

				\textbf{Plankton}  & & & & & & \\

				\rowcolor{LightGray}

			 Baseline  &
				0.8872 & 0.8652 $\pm$ 0.0212 & 0.2584 & 0.2915 $\pm$ 0.0240 & -0.6287 & -0.5737 $\pm$ 0.0444 &  - & - & 15 \\
				 + CE$^-1$  &
				\textbf{0.8896} & \textbf{0.8803 $\pm$ 0.0060} & \textbf{0.2540} & \textbf{0.2690 $\pm$ 0.0098} & \textbf{-0.6356} & \textbf{-0.6113 $\pm$ 0.0154} &  - & - & 12 \\
				 + CC  ($p_A = 0.6$)&
				\textbf{0.8919} & \textbf{0.9128 $\pm$ 0.0427} & 0.4085 & 0.7702 $\pm$ 0.1630 & -0.4833 & -0.1426 $\pm$ 0.1375 & 0.6242 & 0.5927 $\pm$ 0.0127 & 12 \\
				 + DC3 ($p_A = 0.44$)   &
				0.8625 & \textbf{0.9049 $\pm$ 0.0340} & \textbf{0.2192} & 0.3269 $\pm$ 0.0526 & \textbf{-0.6433} & \textbf{-0.5780 $\pm$ 0.0305} &  0.4365 & 0.4451 $\pm$ 0.0204 & 11 \\
				 + DC3 ($p_A = 0.6$)  &
				\textbf{0.9130} & \textbf{0.8768 $\pm$ 0.0640} & \textbf{0.2484} & 0.3004 $\pm$ 0.0750 & \textbf{-0.6646} & \textbf{-0.5764 $\pm$ 0.0416} & 0.6164 & 0.5893 $\pm$ 0.0202 & 14 \\
				\midrule

				\textbf{Turkey}  & & & & & & \\

				\rowcolor{LightGray}

				Baseline  &
				0.8211 & 0.8213 $\pm$ 0.00s69 & 0.3946 & 0.4428 $\pm$ 0.0209 & -0.4265 & -0.3786 $\pm$ 0.0230 & - & - & 15 \\
				+ CE$^-1$  &
				0.7998 & 0.7998 $\pm$ 0.0000 & \textbf{0.3338} & \textbf{0.3338 $\pm$ 0.0000} & \textbf{-0.4660} & \textbf{-0.4660 $\pm$ 0.0000} &  - & - & 12 \\
				+ CC  ($p_A = 0.6$) &
				\textbf{0.8527} & \textbf{0.8264 $\pm$ 0.0469} & \textbf{0.3400} & \textbf{0.3435 $\pm$ 0.0408} & \textbf{-0.5127} & \textbf{-0.4829 $\pm$ 0.0128} & 0.5837 & 0.5646 $\pm$ 0.0427 & 12 \\
				+ DC3 ($p_A = 0.22$)  &
				0.7998 & 0.7998 $\pm$ 0.0000 & \textbf{0.1675} & \textbf{0.2252 $\pm$ 0.0646} & \textbf{-0.6322} & \textbf{-0.5746 $\pm$ 0.0646} &  0.5000 & 0.3674 $\pm$ 0.2054 & 4 \\
				+ DC3  ($p_A = 0.6$)  &
				\textbf{0.8743} & \textbf{0.8432 $\pm$ 0.0350} & \textbf{0.2333} & \textbf{0.3270 $\pm$ 0.0692} & \textbf{-0.6410} & \textbf{-0.5162 $\pm$ 0.0643} &  0.8093 & 0.6387 $\pm$ 0.2354 & 12 \\

					\midrule

		\end{tabular}
	}

\end{table*}
}

\newcommand{\tblAblationPi}{
\begin{table*}[h]

	\caption{
		Complete ablation results for Pi-Model \cite{temporal-ensembling} --
		The vanilla algorithm is highlighted in light grey.
		The row below that extend the algorithm with CE$^-1$ \cite{foc}, Clustering \& Classification (CC) or both (DC3).
		Better results in comparison to the vanilla algorithm are marked bold
	}
	\label{tbl:ablation-pi}

	\resizebox{\textwidth}{!}{
			\begin{tabular}{l c c c c c c c c c c}
				\toprule
			& \multicolumn{2}{c}{F1}  & \multicolumn{2}{c}{\disx}  & \multicolumn{2}{c}{\diffx} & \multicolumn{2}{c}{ Ambiguous } & \# Runs\\

\cmidrule(r){2-3} \cmidrule(r){4-5} \cmidrule(r){6-7}  \cmidrule(r){8-9}

				& best & mean $\pm$ std  &  best & mean $\pm$ std   &  best & mean $\pm$ std &  best  & mean $\pm$ std  & \\

			\midrule

			\textbf{CIFAR-10H}  & & & & & & \\

			\rowcolor{LightGray}

			Baseline  &
			0.7153 & 0.7153 $\pm$ 0.0005 & 0.4913 & 0.5008 $\pm$ 0.0085 & -0.2240 & -0.2145 $\pm$ 0.0087 &  - &- & 3 \\

			 + CE$^-1$  &
			\textbf{0.7255} & \textbf{0.7163 $\pm$ 0.0109 }& 0.4917 & \textbf{0.4988 $\pm$ 0.0138} &\textbf{-0.2337} & \textbf{-0.2175 $\pm$ 0.0244} &  - &- & 3 \\

			+ CC ($p_A = 0.6$)&
			\textbf{0.8420} & 0.6991 $\pm$ 0.1238 & 0.8670 & 0.8823 $\pm$ 0.0133 & 0.0250 & 0.1832 $\pm$ 0.1371 &  0.6120 & 0.5728 $\pm$ 0.0349 & 3 \\

			+ DC3 ($p_A = 0.32$) &
		\textbf{0.7291} &\textbf{ 0.7038 $\pm$ 0.0506} & \textbf{0.3528} & \textbf{0.3564 $\pm$ 0.0849} & \textbf{-0.3764} & \textbf{-0.3474 $\pm$ 0.0466} &  0.3230 & 0.3193 $\pm$ 0.0095 & 3 \\

			+ DC3 ($p_A = 0.6$)  &
		\textbf{0.8305}& \textbf{0.8352 $\pm$ 0.0143} &\textbf{0.4340} & \textbf{0.4680 $\pm$ 0.0311} &\textbf{ -0.3965} &\textbf{ -0.3672 $\pm$ 0.0258} & 0.6125 & 0.6122 $\pm$ 0.0035 & 3 \\

			\midrule

			\textbf{Plankton}  & & & & & & \\

			\rowcolor{LightGray}

			Baseline  &
			0.8757 & 0.8747 $\pm$ 0.0024 & 0.2843 & 0.2840 $\pm$ 0.0019 & -0.5914 & -0.5907 $\pm$ 0.0007 &  - &- & 3 \\

			+ CE$^-1$  &
			\textbf{0.8826} &\textbf{ 0.8783 $\pm$ 0.0043} & \textbf{0.2700} & \textbf{0.2745 $\pm$ 0.0084} & \textbf{-0.6126} & \textbf{-0.6038 $\pm$ 0.0122 }& - &- & 3 \\

			+ CC ($p_A = 0.6$)  &
			0.8027 & \textbf{0.8899 $\pm$ 0.0773} & 0.4346 & 0.7044 $\pm$ 0.2336 & -0.3681 & -0.1855 $\pm$ 0.1594 & 0.5708 & 0.5869 $\pm$ 0.0159 & 3 \\
			+ CC ($p_A = 0.6$)  &
			\textbf{0.9260} &\textbf{ 0.9260 $\pm$ 0.0028} & 0.3411 & 0.3633 $\pm$ 0.0216 & -0.5850 & -0.5627 $\pm$ 0.0231 &  0.4597 & 0.4585 $\pm$ 0.0064 & 3 \\

				+ DC3 ($p_A = 0.6$)  &
			0.7979 & 0.8561 $\pm$ 0.0910 &\textbf{ 0.1908} &\textbf{ 0.2613 $\pm$ 0.0960} & \textbf{-0.6071} & \textbf{-0.5948 $\pm$ 0.0108} &  0.5782 & 0.5829 $\pm$ 0.0042 & 3 \\

			\midrule

			\textbf{Turkey}  & & & & & & \\

			\rowcolor{LightGray}

    		Baseline &
			0.8211 & 0.8172 $\pm$ 0.0038 & 0.3946 & 0.4253 $\pm$ 0.0398 & -0.4265 & -0.3919 $\pm$ 0.0410 &  - &- & 3 \\

			+ CE$^-1$ &
			0.7998 & 0.7998 $\pm$ 0.0000 &\textbf{ 0.3338} & \textbf{0.3338 $\pm$ 0.0000} &\textbf{ -0.4660} & \textbf{-0.4660 $\pm$ 0.0000} & - &- & 3 \\

			+ CC ($p_A = 0.6$) &
			0.7828 & 0.7987 $\pm$ 0.0166 & \textbf{0.3071} & \textbf{0.3266 $\pm$ 0.0192 }& \textbf{-0.4758} &\textbf{ -0.4722 $\pm$ 0.0031} & 0.6025 & 0.5800 $\pm$ 0.0289 & 3 \\

				+ DC3 ($p_A = 0.22$)   &
			- & - & - &- & - & - & - & - & 0 \\

				+ DC3 ($p_A = 0.6$)  &
			\textbf{0.8743} &\textbf{ 0.8768 $\pm$ 0.0035} &\textbf{ 0.2333} & \textbf{0.3071 $\pm$ 0.1044} &\textbf{ -0.6410} & \textbf{-0.5697 $\pm$ 0.1009} & 0.8093 & 0.9047 $\pm$ 0.1348 & 2 \\

			\midrule

			\textbf{Mice Bone}  & & & & & & \\

			\rowcolor{LightGray}

			Baseline  &
			0.6815 & 0.6673 $\pm$ 0.0123 & 0.5411 & 0.5510 $\pm$ 0.0150 & -0.1403 & -0.1164 $\pm$ 0.0237 & - &- & 3 \\

				+ DC3 ($p_A = 0.6$)  &
			\textbf{0.8801} & \textbf{0.8575 $\pm$ 0.0228} & \textbf{0.3099} & \textbf{0.4348 $\pm$ 0.1424} & \textbf{-0.5702} & \textbf{-0.4227 $\pm$ 0.1649} &  0.5621 & 0.5266 $\pm$ 0.0307 & 3 \\

			\midrule

			\textbf{STL-10}  & & & & & & \\

			\rowcolor{LightGray}

			Baseline  &
			0.8256 & 0.8013 $\pm$ 0.0169 & - & - & -0.8256 & -0.8013 $\pm$ 0.0169 &  - &- & 3 \\

				+ DC3 ($p_A = 0.6$)  &
			\textbf{0.8954} & 0.7951 $\pm$ 0.0856 & - & - & \textbf{-0.8954} & -0.7951 $\pm$ 0.0856 &  0.5823 & 0.3425 $\pm$ 0.3128 & 3 \\

		\end{tabular}
	}

\end{table*}
}

\newcommand{\tblAblationCE}{
\begin{table*}[h]

	\caption{
		Complete ablation results for Cross-Entropy --
		The vanilla algorithm is highlighted in light grey.
		The row below that extend the algorithm with CE$^-1$ \cite{foc}, Clustering \& Classification (CC) or both (DC3).
		Better results in comparison to the vanilla algorithm are marked bold
	}
	\label{tbl:ablation-ce}

\resizebox{\textwidth}{!}{
			\begin{tabular}{l c c c c c c c c c c}
				\toprule
			& \multicolumn{2}{c}{F1}  & \multicolumn{2}{c}{\disx}  & \multicolumn{2}{c}{\diffx} & \multicolumn{2}{c}{ Ambiguous } & \# Runs\\

\cmidrule(r){2-3} \cmidrule(r){4-5} \cmidrule(r){6-7}  \cmidrule(r){8-9}

				& best & mean $\pm$ std  &  best & mean $\pm$ std   &  best & mean $\pm$ std &  best  & mean $\pm$ std  & \\

			\midrule

			\textbf{CIFAR-10H}  & & & & & & \\

			\rowcolor{LightGray}

			Baseline  &
			0.6771 & 0.6704 $\pm$ 0.0062 & 0.5580 & 0.5627 $\pm$ 0.0047 & -0.1191 & -0.1077 $\pm$ 0.0099 & - & - & 3 \\

			+ CE$^-1$  &
			\textbf{0.7383} &\textbf{ 0.7329 $\pm$ 0.0078} & \textbf{0.4692} & \textbf{0.4712 $\pm$ 0.0044} &\textbf{ -0.2691} &\textbf{ -0.2618 $\pm$ 0.0120} &  - & -  & 3 \\

			+ CC ($p_A = 0.6$)  &
			\textbf{0.8570} &\textbf{ 0.8518 $\pm$ 0.0049} & 0.8666 & 0.8662 $\pm$ 0.0005 & 0.0096 & 0.0144 $\pm$ 0.0049 & 0.6240 & 0.6197 $\pm$ 0.0045 & 3 \\

				+ DC3 ($p_A = 0.32$) &
			0.6656 & 0.6656 $\pm$ 0.0055 & 0.\textbf{2155} & \textbf{0.2498 $\pm$ 0.0399} &\textbf{ -0.4501} &\textbf{ -0.4158 $\pm$ 0.0364} & 0.2910 & 0.2960 $\pm$ 0.0044 & 3 \\

				+ DC3 ($p_A = 0.6$)  &
			\textbf{0.7827} & 0.6474 $\pm$ 0.1178 &\textbf{ 0.5452} &\textbf{ 0.5096 $\pm$ 0.0382} & \textbf{-0.2375} & \textbf{-0.1378 $\pm$ 0.0870} &  0.6240 & 0.5775 $\pm$ 0.0404 & 3 \\

			\midrule

			\textbf{Plantkon}  & & & & & & \\

			\rowcolor{LightGray}

			Baseline  &
			0.8671 & 0.8632 $\pm$ 0.0034 & 0.3045 & 0.3057 $\pm$ 0.0044 & -0.5626 & -0.5574 $\pm$ 0.0062 &  - & -  & 3 \\

			+ CE$^-1$  &
			\textbf{0.8896} &\textbf{ 0.8880 $\pm$ 0.0023} & \textbf{0.2540} & \textbf{0.2602 $\pm$ 0.0087} & \textbf{-0.6356} &\textbf{ -0.6278 $\pm$ 0.0110} &  - & - & 2 \\

			+ CC ($p_A = 0.6$)  &
			\textbf{0.9596} & \textbf{0.9221 $\pm$ 0.0337} & 0.8321 & 0.8419 $\pm$ 0.0085 & -0.1274 & -0.0802 $\pm$ 0.0422 & 0.5908 & 0.5949 $\pm$ 0.0036 & 3 \\

				+ DC3 ($p_A = 0.44$)  &
			\textbf{0.8625} &\textbf{ 0.9148 $\pm$ 0.0461} & \textbf{0.2192} & \textbf{0.3090 $\pm$ 0.0810} & \textbf{-0.6433} & \textbf{-0.6058 $\pm$ 0.0354} & 0.4365 & 0.4511 $\pm$ 0.0127 & 3 \\

				+ DC3 ($p_A = 0.6$)  &
			0.7824 & \textbf{0.8942 $\pm$ 0.0974} & \textbf{0.2341 }& \textbf{0.3580 $\pm$ 0.1074} & -0.5484 & -0.5361 $\pm$ 0.0155 &  0.5615 & 0.5873 $\pm$ 0.0241 & 3 \\

			\midrule

			\textbf{Turkey}  & & & & & & \\

			\rowcolor{LightGray}

			Baseline  &
	0.8384 & 0.8307 $\pm$ 0.0071 & 0.4298 & 0.4357 $\pm$ 0.0058 & -0.4086 & -0.3951 $\pm$ 0.0117 & - & -  & 3 \\

		+ CE$^-1$  &
	0.7998 & 0.7998 $\pm$ 0.0000 & \textbf{0.3338} &\textbf{ 0.3338 $\pm$ 0.0000} &\textbf{ -0.4660} &\textbf{ -0.4660 $\pm$ 0.0000} & - & -  & 3 \\

		+ CC ($p_A = 0.6$)  &
\textbf{	0.8452} & 0.8033 $\pm$ 0.0550 &\textbf{ 0.3565} & \textbf{0.3213 $\pm$ 0.0542} & \textbf{-0.4887} &\textbf{ -0.4820 $\pm$ 0.0068} &  0.5156 & 0.5409 $\pm$ 0.0295 & 3 \\

		+ DC3 ($p_A = 0.22$)  &
	0.7998 & 0.7998 $\pm$ nan & \textbf{0.2705} & \textbf{0.2705 $\pm$ nan} &\textbf{ -0.5293} & \textbf{-0.5293 $\pm$ nan} & 0.1087 & 0.1087 $\pm$ nan & 1 \\

		+ DC3 ($p_A = 0.6$)  &
	\textbf{0.8579} & 0.8195 $\pm$ 0.0624 & \textbf{0.2764} & \textbf{0.2653 $\pm$ 0.0627} & \textbf{-0.5814} &\textbf{ -0.5543 $\pm$ 0.0252} &  0.5694 & 0.5841 $\pm$ 0.0776 & 3 \\

			\midrule

			\textbf{Mice Bone}  & & & & & & \\

			\rowcolor{LightGray}

			Baseline &
			0.6955 & 0.6753 $\pm$ 0.0198 & 0.5475 & 0.5667 $\pm$ 0.0166 & -0.1479 & -0.1086 $\pm$ 0.0353 & - & -  & 3 \\

				+ DC3 ($p_A = 0.6$)  &
			\textbf{0.9388} & \textbf{0.7373 $\pm$ 0.3046} &\textbf{ 0.3658 }& \textbf{0.3018 $\pm$ 0.1005} & \textbf{-0.5730} & \textbf{-0.4355 $\pm$ 0.2041 } & 0.5680 & 0.5365 $\pm$ 0.0448 & 3 \\

		\midrule

		\textbf{STL-10}  & & & & & & \\

		\rowcolor{LightGray}

		Baseline  &
		0.8048 & 0.7918 $\pm$ 0.0125 & - & - & -0.8048 & -0.7918 $\pm$ 0.0125 & - & - & 3 \\

			+ DC3 ($p_A = 0.6$)  &
	\textbf{	0.8845} & \textbf{0.8671 $\pm$ 0.0166} & - & -& \textbf{-0.8845} &\textbf{ -0.8671} $\pm$ 0.0166 & 0.5919 & 0.4727 $\pm$ 0.2643 & 3 \\
		\end{tabular}
	}

\end{table*}
}

\newcommand{\tblAblationMean}{
\begin{table*}[h]

		\caption{
	Complete ablation results for Mean-Teacher \cite{mean-teacher} --
		The vanilla algorithm is highlighted in light grey.
		The row below that extend the algorithm with CE$^-1$ \cite{foc}, Clustering \& Classification (CC) or both (DC3).
		Better results in comparison to the vanilla algorithm are marked bold
	}
	\label{tbl:ablation-mean}

	\resizebox{\textwidth}{!}{
			\begin{tabular}{l c c c c c c c c c c}
				\toprule
			& \multicolumn{2}{c}{F1}  & \multicolumn{2}{c}{\disx}  & \multicolumn{2}{c}{\diffx} & \multicolumn{2}{c}{ Ambiguous } & \# Runs\\

\cmidrule(r){2-3} \cmidrule(r){4-5} \cmidrule(r){6-7}  \cmidrule(r){8-9}

				& best & mean $\pm$ std  &  best & mean $\pm$ std   &  best & mean $\pm$ std &  best  & mean $\pm$ std  & \\

			\midrule

			\textbf{CIFAR-10H}  & & & & & & \\

			\rowcolor{LightGray}

			Baseline  &
			0.7353 & 0.7280 $\pm$ 0.0065 & 0.4693 & 0.4807 $\pm$ 0.0106 & -0.2659 & -0.2473 $\pm$ 0.0171 & - &- & 3 \\

			+ CE$^-1$  &
			\textbf{0.7360} &\textbf{ 0.7297 $\pm$ 0.0054} & \textbf{0.4747} &\textbf{ 0.4753 $\pm$ 0.0011} & -0.2613 &\textbf{ -0.2544 $\pm$ 0.0061} & - &- & 3 \\

			+ CC ($p_A = 0.6$)  &
			\textbf{0.8565} &\textbf{ 0.7791 $\pm$ 0.1243} & 0.8657 & 0.8747 $\pm$ 0.0153 & 0.0092 & 0.0956 $\pm$ 0.1396 &  0.6145 & 0.5962 $\pm$ 0.0375 & 3 \\

				+ DC3 ($p_A = 0.32$)  &
			0.6614 & 0.7243 $\pm$ 0.0554 &\textbf{ 0.3197 }& \textbf{0.4615 $\pm$ 0.1272 }&\textbf{ -0.3417} & \textbf{-0.2628 $\pm$ 0.0805} & 0.2910 & 0.3070 $\pm$ 0.0151 & 3 \\

				+ DC3 ($p_A = 0.6$)  &
			\textbf{0.8513} & 0.7066 $\pm$ 0.1260 & 0.5244 &\textbf{ 0.4328 $\pm$ 0.0839} & \textbf{-0.3269} & \textbf{-0.2738 $\pm$ 0.0610} &  0.6145 & 0.5757 $\pm$ 0.0336 & 3 \\

			\midrule

			\textbf{Plankton}  & & & & & & \\

			\rowcolor{LightGray}

			Baseline  &
			0.8872 & 0.8828 $\pm$ 0.0044 & 0.2584 & 0.2620 $\pm$ 0.0037 & -0.6287 & -0.6208 $\pm$ 0.0080 &  - &- & 3 \\

			+ CE$^-1$  &
			0.8846 & 0.8821 $\pm$ 0.0035 & \textbf{0.2568} & 0.2623 $\pm$ 0.0050 & -0.6278 & -0.6198 $\pm$ 0.0082 &  - &- & 3 \\

			+ CC ($p_A = 0.6$) &
			\textbf{0.9645} &\textbf{ 0.9345 $\pm$ 0.0260} & 0.8303 & 0.8377 $\pm$ 0.0065 & -0.1342 & -0.0968 $\pm$ 0.0324 &  0.5928 & 0.5898 $\pm$ 0.0029 & 3 \\
				+ DC3 ($p_A = 0.44$)  &
			0.8690 & 0.8634 $\pm$ 0.0080 & 0.2753 & 0.2966 $\pm$ 0.0301 & -0.5937 & -0.5667 $\pm$ 0.0381 &  0.4064 & 0.4098 $\pm$ 0.0049 & 2 \\
				+ DC3 ($p_A = 0.6$)  &
			\textbf{0.9130} &\textbf{ 0.9056 $\pm$ 0.0087} & \textbf{0.2484} & 0.2699 $\pm$ 0.0267 & \textbf{-0.6646} &\textbf{ -0.6357 $\pm$ 0.0282} &  0.6164 & 0.6124 $\pm$ 0.0143 & 3 \\

			\midrule

			\textbf{Turkey}  & & & & & & \\

			\rowcolor{LightGray}

			Baseline  &
			0.8182 & 0.8158 $\pm$ 0.0049 & 0.4512 & 0.4579 $\pm$ 0.0078 & -0.3670 & -0.3579 $\pm$ 0.0079 &  - &- & 3 \\

			+ CE$^-1$  &
			0.7998 & 0.7998 $\pm$  0.0000 & \textbf{0.3338} & \textbf{0.3338 $\pm$  0.0000} & \textbf{-0.4660 }& \textbf{-0.4660 $\pm$ 0.0000 }&  - &- & 3 \\

			+ CC ($p_A = 0.6$)   &
			\textbf{0.8527} &\textbf{ 0.8829 $\pm$ 0.0295} &\textbf{ 0.3400} & \textbf{0.3816 $\pm$ 0.0384} &\textbf{ -0.5127} & \textbf{-0.5013 $\pm$ 0.0098} & 0.5837 & 0.5428 $\pm$ 0.0378 & 3 \\
				+ DC3 ($p_A = 0.22$)  &
			0.7998 & 0.7998 $\pm$ nan &\textbf{ 0.1719} &\textbf{ 0.1719 $\pm$ nan} & \textbf{-0.6278} & \textbf{-0.6278 $\pm$ nan }& 0.5252 & 0.4748 $\pm$ nan & 1 \\

				+ DC3 ($p_A = 0.6$)  &
			\textbf{0.8645} & \textbf{0.8639 $\pm$ 0.0008} &\textbf{ 0.3392} & \textbf{0.3439 $\pm$ 0.0067} &\textbf{ -0.5253} & \textbf{-0.5200 $\pm$ 0.0075} &  0.7691 & 0.7970 $\pm$ 0.0394 & 2 \\

			\midrule

			\textbf{Mice Bone}  & & & & & & \\

			\rowcolor{LightGray}

			Baseline  &
			0.6641 & 0.6688 $\pm$ 0.0217 & 0.4883 & 0.5209 $\pm$ 0.0284 & -0.1758 & -0.1479 $\pm$ 0.0301 & - &- & 3 \\
				+ DC3 ($p_A = 0.6$)  &
			\textbf{0.8984} &\textbf{ 0.8940 $\pm$ 0.0124} & \textbf{0.3511 }& \textbf{0.4300 $\pm$ 0.0887 }& \textbf{-0.5473} & \textbf{-0.4641 $\pm$ 0.0848} &  0.5266 & 0.5444 $\pm$ 0.0178 & 3 \\

	\midrule

			\textbf{STL-10}  & & & & & & \\

			\rowcolor{LightGray}

			Baseline  &
			0.8067 & 0.7863 $\pm$ 0.0173 & - & - & -0.8067 & -0.7863 $\pm$ 0.0173 &  - &- & 3 \\

				+ DC3 ($p_A = 0.6$)  &
			\textbf{0.8928} & \textbf{0.8751 $\pm$ 0.0188} & - & - & \textbf{-0.8928} & \textbf{-0.8751 $\pm$ 0.0188} & 0.5897 & 0.4732 $\pm$ 0.2646 & 3 \\

		\end{tabular}
	}

\end{table*}
}

\newcommand{\tblAblationPseudo}{
\begin{table*}[h]

		\caption{
		Complete ablation results for Pseudo-Label \cite{pseudolabel} --
		The vanilla algorithm is highlighted in light grey.
		The row below that extend the algorithm with CE$^-1$ \cite{foc}, Clustering \& Classification (CC) or both (DC3).
		Better results in comparison to the vanilla algorithm are marked bold
	}
	\label{tbl:ablation-pseudo}

	\resizebox{\textwidth}{!}{
			\begin{tabular}{l c c c c c c c c c c}
				\toprule
			& \multicolumn{2}{c}{F1}  & \multicolumn{2}{c}{\disx}  & \multicolumn{2}{c}{\diffx} & \multicolumn{2}{c}{ Ambiguous } & \# Runs\\

\cmidrule(r){2-3} \cmidrule(r){4-5} \cmidrule(r){6-7}  \cmidrule(r){8-9}

				& best & mean $\pm$ std  &  best & mean $\pm$ std   &  best & mean $\pm$ std &  best  & mean $\pm$ std  & \\

			\midrule

			\textbf{CIFAR-10H}  & & & & & & \\

			\rowcolor{LightGray}

			Baseline  &
			0.6970 & 0.6914 $\pm$ 0.0057 & 0.5330 & 0.5359 $\pm$ 0.0050 & -0.1640 & -0.1554 $\pm$ 0.0103 & - &- & 3 \\

		+ CE$^-1$ &
			\textbf{0.7054} & \textbf{0.6977 $\pm$ 0.0108} & \textbf{0.5194} & \textbf{0.5264 $\pm$ 0.0113} & \textbf{-0.1860} & \textbf{-0.1713 $\pm$ 0.0221} &  - &- & 3 \\

			+ CC ($p_A = 0.6$)   &
			\textbf{0.8265} & 0.6583 $\pm$ 0.1457 & 0.8670 & 0.8838 $\pm$ 0.0147 & 0.0404 & 0.2255 $\pm$ 0.1603 & 0.6190 & 0.5803 $\pm$ 0.0337 & 3 \\

				+ DC3 ($p_A = 0.32$)  &
			0.6236 & \textbf{0.6941 $\pm$ 0.0611} & \textbf{0.2382} & \textbf{0.4058 $\pm$ 0.1464} &\textbf{-0.3854} & \textbf{-0.2883 $\pm$ 0.0870} &  0.3245 & 0.3237 $\pm$ 0.0063 & 3 \\

				+ DC3 ($p_A = 0.6$)  &
			\textbf{0.8374} & \textbf{0.7448 $\pm$ 0.1440} & \textbf{0.5132} & \textbf{0.4854 $\pm$ 0.0967} & \textbf{-0.3242} & \textbf{-0.2594 $\pm$ 0.0618} &  0.6100 & 0.5957 $\pm$ 0.0311 & 3 \\

			\midrule

			\textbf{Plankton}  & & & & & & \\

			\rowcolor{LightGray}

			Baseline  &
			0.8762 & 0.8730 $\pm$ 0.0045 & 0.2742 & 0.2821 $\pm$ 0.0098 & -0.6020 & -0.5908 $\pm$ 0.0143 &  - &- & 3 \\

			+ CE$^-1$  &
			\textbf{0.8737} & 0.8727 $\pm$ 0.0015 & 0.2788 & \textbf{0.2794 $\pm$ 0.0009} & -0.5949 & \textbf{-0.5932 $\pm$ 0.0024} &  - &- & 2 \\

			+ CC ($p_A = 0.6$)   &
			\textbf{0.8919} & \textbf{0.9046 $\pm$ 0.0220} & 0.4085 & 0.6967 $\pm$ 0.2497 & -0.4833 & -0.2078 $\pm$ 0.2400 & 0.6242 & 0.5991 $\pm$ 0.0221 & 3 \\

				+ DC3 ($p_A = 0.44$)  &
			0.8640 & \textbf{0.9016 $\pm$ 0.0327} & \textbf{0.2661} & 0.3285 $\pm$ 0.0545 & -0.5979 & -0.5731 $\pm$ 0.0217 & 0.4304 & 0.4491 $\pm$ 0.0167 & 3 \\

				+ DC3 ($p_A = 0.6$)  &
			\textbf{0.8931} & 0.8539 $\pm$ 0.0555 & 0.3176 & 0.2844 $\pm$ 0.0469 & -0.5755 & -0.5695 $\pm$ 0.0085 & 0.5945 & 0.5843 $\pm$ 0.0144 & 2 \\

			\midrule

			\textbf{Turkey}  & & & & & & \\

			\rowcolor{LightGray}

			Baseline  &
			0.8237 & 0.8245 $\pm$ 0.0012 & 0.4488 & 0.4527 $\pm$ 0.0056 & -0.3749 & -0.3718 $\pm$ 0.0044 &  - &- & 2 \\

		+ CE$^-1$ &
			0.7998 & 0.7998 $\pm$ 0.0000 & \textbf{0.3338} & \textbf{0.3338 $\pm$ 0.0000} & \textbf{-0.4660} & \textbf{-0.4660 $\pm$ 0.0000} &  - &- & 3 \\

			+ CC ($p_A = 0.6$)   &
			\textbf{0.8486} & 0.8207 $\pm$ 0.0334 & \textbf{0.3708} & \textbf{0.3445 $\pm$ 0.0319} & \textbf{-0.4778} & \textbf{-0.4762 $\pm$ 0.0016} & 0.6310 & 0.5947 $\pm$ 0.0601 & 3 \\

				+ DC3 ($p_A = 0.22$)  &
			0.7998 & 0.7998 $\pm$ 0.0000 & \textbf{0.1675} & \textbf{0.2291 $\pm$ 0.0871} & \textbf{-0.6322} & \textbf{-0.5706 $\pm$ 0.0871} & 0.5000 & 0.4783 $\pm$ 0.0307 & 2 \\

				+ DC3 ($p_A = 0.6$) &
			\textbf{0.8344} & \textbf{0.8292 $\pm$ 0.0074} & \textbf{0.3504} & \textbf{0.3883 $\pm$ 0.0536} & \textbf{-0.4841} & \textbf{-0.4409 $\pm$ 0.0610} &  0.8560 & 0.5305 $\pm$ 0.4604 & 2 \\

			\midrule

			\textbf{Mice Bone}  & & & & & & \\

			\rowcolor{LightGray}

			Baseline  &
			0.6660 & 0.6524 $\pm$ 0.0124 & 0.5703 & 0.5768 $\pm$ 0.0057 & -0.0957 & -0.0756 $\pm$ 0.0176 & - &- & 3 \\

				+ DC3 ($p_A = 0.6$)  &
			\textbf{0.8658} & \textbf{0.7275 $\pm$ 0.2267} &\textbf{0.3752} & \textbf{0.3465 $\pm$ 0.0978} & \textbf{-0.4906} & \textbf{-0.3810 $\pm$ 0.1363} & 0.5444 & 0.5444 $\pm$ 0.0118 & 3 \\

			\midrule

			\textbf{STL-10}  & & & & & & \\

			\rowcolor{LightGray}

			Baseline  &
			0.8248 & 0.8016 $\pm$ 0.0184 & - & - & -0.8248 & -0.8016 $\pm$ 0.0184 &  - &- & 3 \\
				+ DC3 ($p_A = 0.6$)  &
			\textbf{0.8887} & 0.7903 $\pm$ 0.0854 & - & - & \textbf{-0.8887} & -0.7903 $\pm$ 0.0854 & 0.5921 & 0.4606 $\pm$ 0.2581 & 3 \\

		\end{tabular}
	}

\end{table*}
}

\newcommand{\tblAblationFixmatch}{
\begin{table*}[h]

	\caption{
		Complete ablation results for FixMatch \cite{fixmatch} --
		The vanilla algorithm is highlighted in light grey.
		The row below that extend the algorithm with CE$^-1$ \cite{foc}, Clustering \& Classification (CC) or both (DC3).
		Better results in comparison to the vanilla algorithm are marked bold
	}
	\label{tbl:ablation-fixmatch}

	\resizebox{\textwidth}{!}{
			\begin{tabular}{l c c c c c c c c c c}
				\toprule
			& \multicolumn{2}{c}{F1}  & \multicolumn{2}{c}{\disx}  & \multicolumn{2}{c}{\diffx} & \multicolumn{2}{c}{ Ambiguous } & \# Runs\\

\cmidrule(r){2-3} \cmidrule(r){4-5} \cmidrule(r){6-7}  \cmidrule(r){8-9}

				& best & mean $\pm$ std  &  best & mean $\pm$ std   &  best & mean $\pm$ std &  best  & mean $\pm$ std  & \\

			\midrule

		\textbf{CIFAR-10H}  & & & & & & \\

		\rowcolor{LightGray}

		Baseline  &
		0.7809 & 0.7713 $\pm$ 0.0097 & 0.4199 & 0.4332 $\pm$ 0.0122 & -0.3611 & -0.3381 $\pm$ 0.0218 &  - &- & 3 \\
			+ DC3 ($p_A = 0.6$)  &
	\textbf{0.8309} & 0.7947 $\pm$ 0.0335 & 0.4949 & 0.4746 $\pm$ 0.0190 & -0.3360 & -0.3200 $\pm$ 0.0145 & 0.5805 & 0.5688 $\pm$ 0.0169 & 3 \\
		\midrule

		\textbf{Plankton}  & & & & & & \\

		\rowcolor{LightGray}

		Baseline  &
		0.8581 & 0.8324 $\pm$ 0.0278 & 0.3029 & 0.3237 $\pm$ 0.0231 & -0.5552 & -0.5088 $\pm$ 0.0509 &  - &- & 3 \\
			+ DC3 ($p_A = 0.6$)  &
		\textbf{0.8720} & \textbf{0.8666 $\pm$ 0.0649} & 0.3128 & \textbf{0.3228 $\pm$ 0.0659} & \textbf{-0.5592} & \textbf{-0.5438 $\pm$ 0.0134 }& 0.5770 & 0.5782 $\pm$ 0.0259 & 3 \\
		\midrule

		\textbf{Turkey}  & & & & & & \\

		\rowcolor{LightGray}

		Baseline  &
		0.8214 & 0.8196 $\pm$ 0.0019 & 0.4333 & 0.4455 $\pm$ 0.0121 & -0.3881 & -0.3741 $\pm$ 0.0130 & - &- & 3 \\
			+ DC3 ($p_A = 0.6$)  &
		\textbf{0.8356} & \textbf{0.8401 $\pm$ 0.0146} & \textbf{0.2817} & \textbf{0.3499 $\pm$ 0.0673} & \textbf{-0.5539} & \textbf{-0.4902 $\pm$ 0.0581} &  0.2691 & 0.4827 $\pm$ 0.1856 & 3 \\
		\midrule

		\textbf{STL-10}  & & & & & & \\

		\rowcolor{LightGray}

		Baseline &
		0.8957 & 0.8948 $\pm$ 0.0011 & - & - & -0.8957 & -0.8948 $\pm$ 0.0011 &  - &- & 3 \\
			+ DC3 ($p_A = 0.6$)  &
		\textbf{0.9145} & 0.8690 $\pm$ 0.0440 & - & - & \textbf{-0.9145} & -0.8690 $\pm$ 0.0440 &  0.5746 & 0.5586 $\pm$ 0.0266 & 3 \\

		\end{tabular}
	}

\end{table*}
}

\usepackage{hyperref}

\usepackage{breakcites}
\usepackage{placeins}
\usepackage{listings}
\usepackage{csquotes}
\usepackage{xspace}
\newcommand\certain{{certain}\xspace}
\newcommand\fuzzy{{ambiguous}\xspace}
\newcommand\diff{($d-$F1)}
\newcommand\diffx{($d-$F1)\xspace}
\newcommand\dis{$d$}
\newcommand\disx{$d$\xspace}

\begin{document}
\pagestyle{headings}
\mainmatter
\def\ECCVSubNumber{xxx}  %

\title{A data-centric approach for improving ambiguous labels with combined semi-supervised classification and clustering} %

\titlerunning{Data-centric classification and clustering (DC3)}
\author{Lars Schmarje\inst{1} \and
Monty Santarossa\inst{1} \and
Simon-Martin Schröder\inst{1}\and
Claudius Zelenka\inst{1}\and
Rainer Kiko\inst{2}\and
Jenny Stracke\inst{3}\and
Nina Volkmann\inst{4}\and
Reinhard Koch\inst{1}}
\authorrunning{L. Schmarje et al.}
\institute{MIP, Kiel University, Germany
\email{\{las,msa,sms,czw,rk\}@informatik.uni-kiel.de} \and
LOV, Sorbonne Université, France
\email{rainer.kiko@obs-vlfr.fr} \and
ITW, University of Bonn
\email{jenny.stracke@itw.uni-bonn.de}\and
WING, Unviersity of Veterinary Medicine Hannover
\email{nina.volkmann@tiho-hannover.de}}
\maketitle

\begin{abstract}
Consistently high data quality is essential for the development of novel loss functions and architectures in the field of deep learning. The existence of such data and labels is usually presumed, while acquiring high-quality datasets is still a major issue in many cases. In real-world datasets we often encounter ambiguous labels due to subjective annotations by annotators. In our data-centric approach, we propose a method to relabel such ambiguous labels instead of implementing the handling of this issue in a neural network.  A hard classification is by definition not enough to capture the real-world ambiguity of the data. Therefore, we propose our method "Data-Centric Classification \& Clustering (DC3)" which combines semi-supervised classification and clustering. It automatically estimates the ambiguity of an image and performs a classification or clustering depending on that ambiguity.  DC3 is general in nature so that it can be used in addition to many Semi-Supervised Learning (SSL) algorithms.  On average, this results in a 7.6\% better F1-Score for classifications and 7.9\% lower inner distance of clusters across multiple evaluated SSL algorithms and datasets. Most importantly, we give a proof-of-concept that the classifications and clusterings from DC3 are beneficial as proposals for the manual refinement of such ambiguous labels. Overall, a combination of SSL with our method DC3 can lead to better handling of ambiguous labels during the annotation process.
\footnote{Source code is available at \url{https://github.com/Emprime/dc3}, Datasets available at \url{https://doi.org/10.5281/zenodo.5550916}}
\end{abstract}

\section{Introduction}

\pic{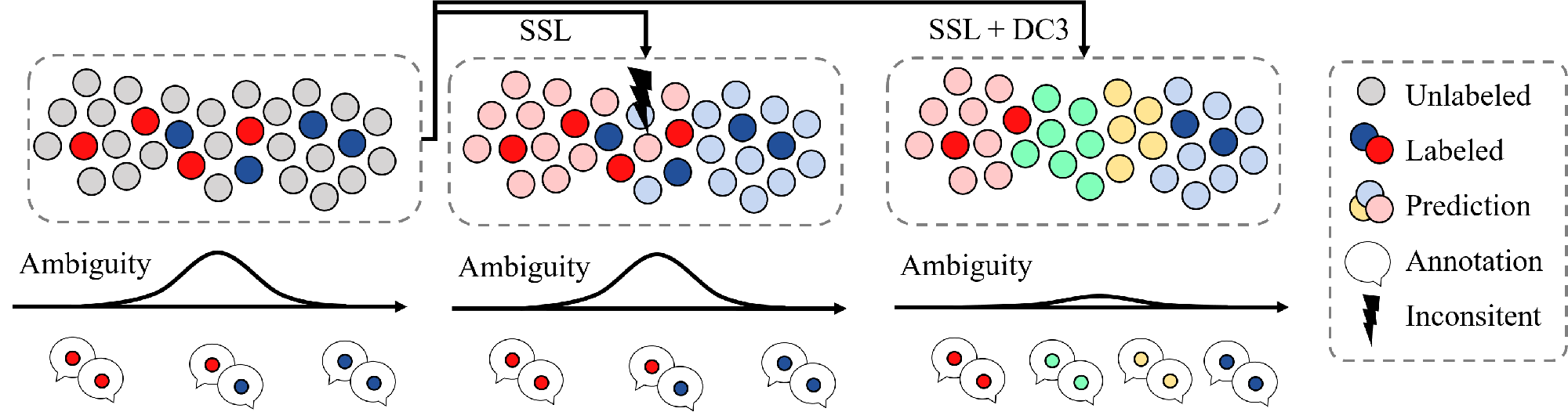}{Benefit of data-centric classification and clustering (DC3) over Semi-Supervised Learning (SSL) -
Real-world datasets often suffer from intra- or interoberserver variability (IIV) during the annotation and thus no clear separation of classes is given as in common benchmark datasets.
Images with a high variability between the annotations have therefore an \fuzzy label.
SSL can be confused by these \fuzzy labels (see lightning bolt) which results in inconsistent predictions.
Our method DC3 can be used in combination with SSL to identify \fuzzy images automatically and cluster them, while classifying the rest as usual.
Therefore, we avoid the ambiguity of the labels during training and generated cluster proposals which can be used to create more consistent labels.
}{0.95}

In recent years, deep learning has been successfully applied to many computer vision problems \cite{maskrcnn,fixmatch,simclrv2,santarossa2022medregnet,MetaPseudo,Damm2021Sofia}.
A main reason for this success were large high-quality datasets which enabled machine learning to incorporate a wide variety of real world patterns \cite{imagenet}.
Many novel loss functions and architectures have been proposed including options to handle imperfect data \cite{imperfect_segmentation,weakMining}.
This model-centric view mostly focuses on trying to deal with issues like label bias \cite{labelbias}, label noise \cite{noisy-labels-comparison} or ambiguous labels \cite{deep-learn-label-distribution} instead of improving the dataset during the annotation process.
Following recent data-centric literature \cite{are_we_done,foc,NotBlackAndWhite}, we therefore investigate in this paper an approach to improve the dataset during the acquisition process.

Specifically, we look at the impact of ambiguous labels introduced due to  \emph{intra- or interobserver variability} (IIV) which describes the variability / inconsistency between the annotations over time or between annotators.
This issue is common when annotating data \cite{cifar10h,NotBlackAndWhite,noisy-labels-comparison,foc,schmarje2019,inter-observer-segmentation,schmarje2022benchmark,mammo-variability,tailception,schmarje2021datacentric,eye-fuzzy,cancergrading,planktonUncertain,grossmann2022beyond}.
The literature names different possible reasons for this variability such as low resolution \cite{cifar10h}, bad quality\cite{ocr-example,schmarje2019}, subjective interpretations of classes \cite{cancergrading,mammo-variability} or mistakes \cite{noisy-labels-comparison,divide-mix}.

We assume that this variability can be modeled for each image with an unknown soft probability distribution $l \in [0,1]^k$ for a classification problem with $k$ classes.
Many previous methods use a hard label instead of a soft label for training and therefore can not model this issue by definition.
We call a label and its corresponding image \emph{\certain} if all annotators would agree on the classification ($l \in \{0,1\}$) and \emph{\fuzzy} if they would disagree ($l \in (0,1)$).
In other words, \fuzzy images are likely to have different annotations due to IIV while certain images do not.
The issue of \fuzzy images is that the unknown distribution $l$ can only be estimated with expensive operations like actually acquiring multiple annotations.
Real-world example images with \certain and \fuzzy labels are given in \autoref{fig:dataset} and detailed definitions are given in \autoref{subsec:definitions}.

The goal of this paper is to introduce a method which provides predictions which are beneficial for improving ambiguous labels via relabeling in a down-stream task.
The quality of ambiguous labels and thus the performance of trained models \cite{are_we_done,relabelImagenet}  can easily be improved with the usage of more annotations.
However, more annotations are associated with a higher cost in the form of human working hours which is not desirable.
Semi-Supervised Learning (SSL) can counter these higher cost because it has shown great potential in reducing the amount of labeled data  to 10\% or even 1\% while maintaining classification performance \cite{fixmatch,selfmatch,simclrv2,barlowtwins,swav} or even boost performance further \cite{noisy-student,MetaPseudo} on already large labeled datasets like ImageNet \cite{imagenet}.

Therefore, we propose Data-Centric Classification \& Clustering (DC3) which can be used in combination with many SSL algorithms to perform a combined semi-supervised classification and clustering.
It simultaneously distinguishes between \fuzzy and \certain images,  classifies the \certain  images and clusters visually similar \fuzzy images.
A graphical summary is provided in \autoref{fig:teaser}.
We will show that this approach leads to better classifications and more compact clusters across multiple semi-supervised algorithms and non-curated datasets.
Furthermore, we give a proof-of-concept that these improvements lead to a greater consistency of labels based on proposals from DC3.
We will discuss different parts of our methods in detail and the limitations.

The key contributions of this paper are:
(1) DC3 allows a SSL algorithm to predict on average a 7.6\% better F1-Score for classifications and a 7.9\% lower inner distance of clustering across multiple algorithms and non-curated datasets.
The hyperparameters of DC3 are fixed across all algorithms and datasets which illustrates the generalizability of our method.
(2) We give a proof-of-concept that these improved predictions can be used to create on average a 2.4 faster and 6.74\% more consistent labels in comparison to the non-extended algorithms and a consensus process.
This leads to higher quality data for further evaluation or model training.
(3) DC3 can be used in combination with many SSL algorithms.
This means DC3 can benefit from novel SSL algorithms in the future because it can be added without a noticeable trade-off in terms of run-time or memory consumption.

\subsection{Related Work}

Our method is mainly related to Data-Centric Machine Learning, Semi-Supervised Learning and Classification \& Clustering.

\paragraph{Data-Centric Machine Learning} aims at improving the data quality rather then improving the model alone\cite{NotBlackAndWhite,data-centric-appraoch}.
The data issues like imperfect, ambiguous or erroneous labels \cite{are_we_done,underwater_uncertainty,relabelImagenet,planktonUncertain,inter-observer-segmentation,dataset_issues,merIssue} are often handeled in a model-centric approach by detecting errors or making the models more robust \cite{song2022noisyLabelsSurvey,robust-semi-supervised,noisy-labels-comparison,surveynoise}.
We want to use the predictions of our model to improve the annotation process and therefore prevent or minimize the quality issues before they need to be handled by special networks.

\paragraph{Semi-Supervised Learning} \cite{semi-supervised-learning} is mainly developed on curated benchmark datasets \cite{imagenet,stl-10,cifar} where the issue of IIV is not considered.
In contrast to other SSL research \cite{simclrv2,barlowtwins,swav,byol,fixmatch}, we are not evaluating on these curated benchmarks but work with new real-world datasets for two reasons.
Firstly, curated datasets do not suffer so much from IIV because they were already cleaned.
Recent research indicates that even these datasets suffer from errors in the labels which negatively impact the performance \cite{cifar10h,are_we_done}.
Secondly, if we want to evaluate the IIV issue, we need an approximation of the variability of the label for each image e.g. in the form of multiple annotations per image.
However, this information is often not provided for current state-of-the-art benchmarks.

\paragraph{Classification\&Clustering} was investigated in detailed \cite{scc,cluster-then-label,scccsr,deep-cluster,Tian2021Divide}.
However, classical low dimensional approaches are difficult to extend to real-world images \cite{scc,cluster-then-label,scccsr}.
and many deep-learning methods use the clustering only as a proxy task before the actual classification \cite{scan,iic,foc} or iterate between classifications and clusters \cite{deep-cluster,Tian2021Divide}.
Smieja et al. are one of the few which use the classification and clustering outputs in parallel in each training step \cite{Smieja2020}.
We want to automatically decide which data should be rather classified or clustered due to their underlying ambiguity.

\section{Method}

Our method  Data-Centric Classification \& Clustering (DC3)  is not a individual method but an extension for most SSL algorithms such as \cite{mixmatch,fixmatch,mean-teacher,pseudolabel,temporal-ensembling}.
We can extend any image classification model with DC3 as long as it is compatible with the definition of an arbitrary SSL algorithm below.

\subsection{Definitions}
\label{subsec:definitions}

We assume that every image $x \in X$ has an unknown soft probability distribution $l \in [0,1]^k$ for a classification problem with $k$ classes.
This assumption is based on two main reasons.
Firstly, inconsistent annotations exist due to subjective opinions from the annotators, e.g. the grading of an illness \cite{cancergrading}.
A hard label $l \in \{0,1\}^k$ could not model such a difference over the complete annotator population.
Secondly, if we look for example at biological processes, there exist images of intermediate transition stages between two classes, e.g. the degeneration of a living underwater organism to dead biomass \cite{foc}.

An image and its corresponding label $l$ are \emph{\fuzzy} if $i,j \in \{1,..,k\}$ exist with $i \neq j, l_i > 0$ and $l_j > 0$.
Otherwise the image and its label are \emph{\certain}.
The ambiguity of a label is $1 - max_{i \in \{1,..,k\}} l_i$.
An image might be \fuzzy because it is actually an intermediate or uncertain combination of different classes as stated above.
For this reason, we view \fuzzy images not just as wrongly assigned images.

A SSL algorithm uses a labeled dataset $X_l$ and an unlabeled dataset $X_u$ for the training of a neural network $\Phi$ with $X = X_l \dot\bigcup X_u$.
For all images $x \in X_l$ a hard label $l$ is available while no label information is available for $x \in X_u$.
The output $p_n(x) := \Phi(x)$ is a probability distribution over the $k$ classes.

\subsection{DC3}

Our method DC3 extends an arbitrary SSL algorithm.
This SSL algorithm passes an image $x$ through the network $\Phi$ and predicts a classification $p_n(x) \in [0,1]^k$.
DC3 calculates two additional outputs without a noticeable impact on training time or memory consumption: a clustering assignment $p_o(x) \in [0,1]^{k'}$ with $k' > k$ and an ambiguity estimation $p_a(x) \in [0,1]$.
The cluster assignment partitions visually similar images in more clusters than classes exist (overclustering with $k' > k$).
The ambiguity estimation is used to determine if a classification ($p_n(x)$) or an (over)clustering ($p_o(x)$) is used as the final output.
If $p_a(x) < 0.5$ the image is predicted as \certain and thus the classification is used.
Otherwise, the image is estimated as \fuzzy and the clustering is used as output.

\pic{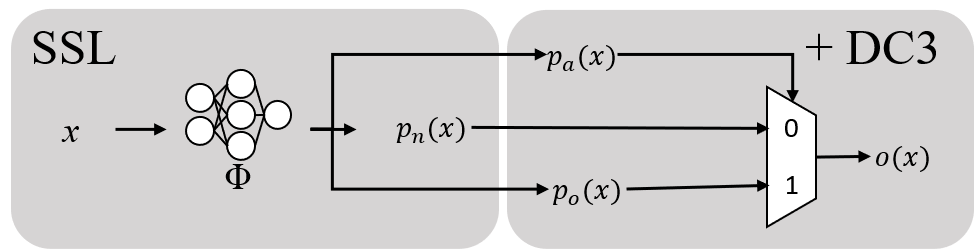}{Our method DC3 and an extended arbitrary SSL method --
The SSL algorithm passes an image $x$ through the network $\Phi$ and outputs a classification $p_n(x)$.
We add two additional outputs: an overclustering $p_o(x)$ and a ambiguity estimation $p_a(x)$.
The ambiguity estimation $p_a(x)$ is used to determine if the classification or the overclustering output is used for our method DC3.
Only some labels are available for the classification output and therefore most images have to be trained completely self-supervised on all outputs.
}{0.95}

A key difference to previous literature \cite{iic,foc,deep-cluster} is that we do not want an additional or only a clustering of all samples.
We want SSL classifications for certain images while these classifications are not well suited for ambiguous images.
On these ambiguous images a clustering is desired without the knowledge of expected cluster assignments.
Moreover, it is not feasible to determine ambiguous images before this classification/clustering and thus we have no ground-truth for this decision as well.
These combinations lead to to three goals which need to be achieved:
1. The underlying SSL training must be possible and not negatively impacted while computing an additional overclustering.
2. No ground-truth is available for the clustering and therefore a degeneration to one or random cluster assignments has to be avoided.
3. The same argument applies to the ambiguity estimation $p_a(x)$ and a balance between certain and ambiguous images is needed.
For this purpose, the network is trained by minimizing the following loss function which benefits from SSL but avoids the described degenerations.
	\begin{equation}
	\label{eq:loss}
	\begin{split}
	L(x) &  = L_{SSL}(x) \cdot [1 - p_a(x)]  + \lambda_{CE^{-1}} L_{CE^{-1}}(x) \cdot [1 - p_a(x)] \\
	& +  \lambda_a L_A(x) +  \lambda_s L_S(x) \cdot p_a(x)
	\end{split}
	\end{equation}
The first three loss terms correspond to the outputs $p_n(x), p_o(x)$ and $p_a(x)$ and the three goals described above, respectively.
The last term ($L_S$) is optional and stabilizes the training.
The $\lambda$ values are weights to balance the impact of each term.
The first loss $L_{SSL}$ is the loss calculated by the original SSL algorithm and is only scaled with $ [1 - p_a(x)]$ to prevent the original SSL training on images the network predicts as \fuzzy.

The second loss $L_{CE^{-1}}$ incentivises visually homogeneous clusters of the images by pushing images from different classes into different clusters.
This loss is needed to prevent a degeneration of the clustering.
A similar loss was used in \cite{foc} but could only be trained on labeled data, with pretrained networks and  several inefficient stabilizing methods like repeating every sample 3-5 times per batch.
We generalized the formula for two input images $x,x'$ of the same mini-batch which should not be of the same class:
	\begin{equation}
	\label{eq:ceinv}
	\begin{split}
	CE^{-1}(p_o(x),p_o(x')) & = - \sum_{c=1}^{k} p_o(x)_c \cdot ln(1-p_o(x')_c)\,.
	\end{split}
	\end{equation}
For the selection of $x,x'$, we use either the ground-truth label $l$ of $x$ if it is available or the Pseudo-Label based on the network prediction $p_n(x)$.
The loss is also scaled with $[1 - p_a(x)]$ because it uses an estimate of the class for an image which could be wrong / ill-suited for \fuzzy images.

The third loss $L_A$ allows the ambiguity estimation.
As stated above, the underlying distribution $l$ is unknown and thus we do not know during training if  $x$ is \fuzzy or \certain.
However, we can expect to know or be given an prior probability $p_A \in [0,1]$ of the expected percentage of \fuzzy images in the total dataset.
We set $p_A$ to a fixed value which balances certain and \fuzzy images and the details are given in \autoref{subsec:details}.
Based on this probability, we can estimate a Pseudo-Label of the ambiguity of each image in a batch during training.
The loss $L_A$ is the binary cross-entropy between the Pseudo-Label $h(x)$ and $p_a(x)$.
The usage of hot-encoded Pseudo-Labels forces the network to make more confident predictions.
The formulation is given below with $i$ the index of the image $x$ inside the given batch when all images inside the batch are sorted in ascending order based on $p_a$.
\begin{equation}
	\label{eq:lf}
	\begin{split}
	L_A(x) & = CE(h(x),p_a(x)) \\
	& = - (1 - h(x)) \cdot ln(p_a(x = 0)) \\
	& - h(x) \cdot ln(p_a(x = 1))  \text{ with }\\
	h(x) &= \begin{cases}
        1 & i \leq \text{batch size} \cdot p_A\\
        0 &\text{else}
        \end{cases}
	\end{split}
\end{equation}

The forth term $L_S$ is the cross-entropy (CE) between $p_o(x)$ and $p_o(x')$ for two differently augmented versions  $x,x'$ of the same image.
This loss is scaled with $p_a(x)$ and incentivises that augmented versions of the same \fuzzy image are in the same output cluster.
We use CE because it indirectly minimizes also the entropy of $p_o(x)$ which leads to sharper predictions.
Many SSL algorithms already use a differently augmented version $x'$ of $x$ as secondary input \cite{mixmatch,fixmatch,mean-teacher,temporal-ensembling,iic} which allows an easy computation. Otherwise, the forth term is not calculated and treated as zero.

It is important to note that only the proposed combination of the individual parts leads to a successful training of all desired outputs.
We show in \autoref{subsec:ablation} that the combined clustering and classification (CC) based on $p_a(x)$ and the loss $L_{CE^{-1}}$ are the two essential parts to DC3.

\section{Experiments}
\subsection{Datasets}

A main contribution of this paper is that our method can be applied to many SSL algorithms across different real-world \fuzzy datasets without major changes.
While many datasets \cite{cifar10h,noisy-labels-comparison,foc,schmarje2019,inter-observer-segmentation,mammo-variability,tailception,eye-fuzzy,cancergrading,planktonUncertain} suffer from annotation variability, we do not know the unknown underlying distribution $l$ to evaluate the ambiguity or any related metrics.
We can approximate $l$ with the average over multiple annotations from humans.
An annotation is the  hard coded guess $a = (a_1,...,a_k) \in \{0,1\}^k$ of the class for an image from a human  with exactly one $i' \in \{1,...,k\} : a_i' = 1$ and for all $j \in \{1,...,k\} \setminus \{i'\} : a_j = 0$.
We assume that the approximation $\hat l$ as the average of $n$ annotations is identical to the unknown distribution $l$ for $n \longrightarrow \infty$.
This leaves the issue that we need multiple annotations per image for a dataset with \fuzzy labels which are often not available.
However, all datasets summarized in \autoref{tbl:datasets} have multiple annotations and thus allow the approximation of $\hat l$.
Nine visual examples for all datasets are given in \autoref{fig:dataset} and the datasets are shortly introduced below.

The \emph{Plankton} dataset was introduced in \cite{foc}.
The dataset contains 10 plankton classes and has multiple labels per image due to the help of citizen scientists.
In contrast to  \cite{foc}, we include \fuzzy images in the training and validation set and do not enforce a class balance which results in a slightly different data split as shown in \autoref{fig:dataset}.

The \emph{Turkey} dataset was used in \cite{turkey_dataset,volkmann_turkeys}.
The dataset contains cropped images of potential injuries which were separately annotated by three experts as not injured or injured. %

The \emph{Mice Bone} dataset is based on the raw data which was published in \cite{schmarje2019}.
The raw data are 3D scans from collagen fibers in mice bones.
The three proposed classes are similar and dissimilar collagen fiber orientations and not relevant regions due to noise or background.
We used the given segmentations to cut image regions from the original 2D image slices which mainly consist of one class.
We generated \fuzzy GT labels on 10\% of the generated images by averaging over three own classifications from an expert.

The \emph{CIFAR-10H} \cite{cifar10h} dataset provides multiple annotations for the test set of CIFAR-10\cite{cifar}.
This dataset is interesting because it illustrates that even the hard labels from benchmark datasets like CIFAR-10 are based on soft labels due to IIV.

\tblDatasets

\picFour{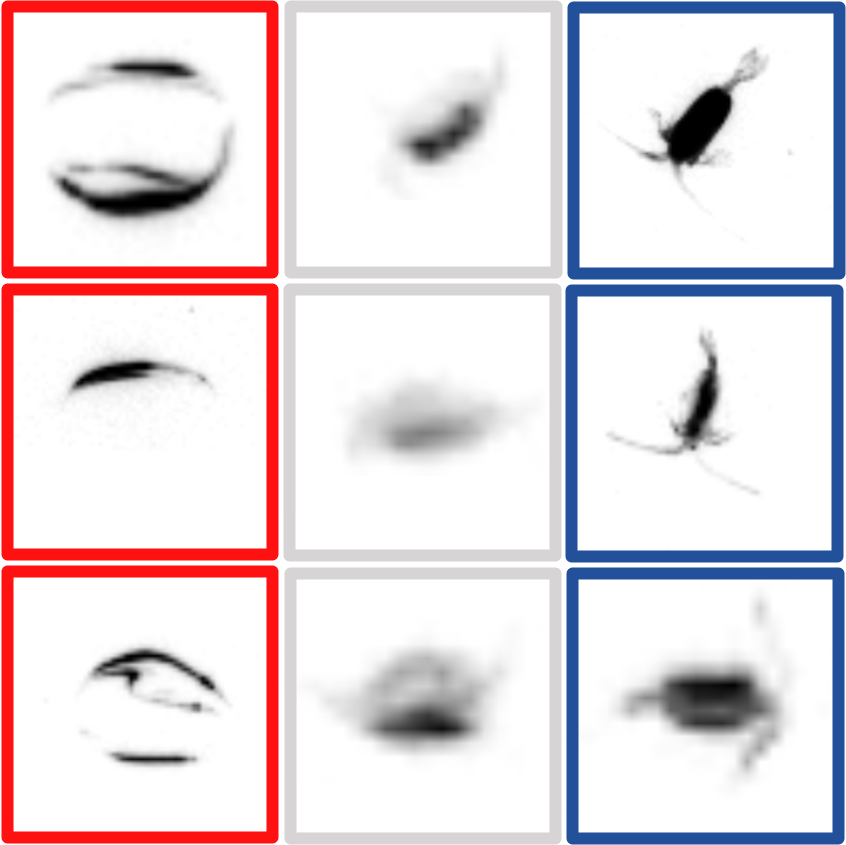}{Plankton \cite{foc}}{Turkey \cite{turkey_dataset}}{Mice Bone \cite{schmarje2019}}{CIFAR-10H \cite{cifar10h}}{Example images for the \fuzzy real-world datasets --
All datasets have \certain  images (red \& blue) and \fuzzy images between these classes (grey). The classes are Bubble \& Copepod, Not Injured \& Injured, Similar \& Dissimilar Orientations and Dog \& Cat respectively.
}

For all datasets, we split our images $X$ into a labeled $X_l$ and unlabeled $X_u$  training set.
We keep additional images as a validation subset.
On $X_l$, we use for each image a random hard label sampled from the corresponding $\hat l$.
This simulates that we only have a noisy approximation of the true ground truth label $l$.
On $X_u$, we can only use the image information and not any label information.
The validation data is used to compare the trained networks and to detect issues like overfitting.

As stated above the approximation of $\hat l$ is only possible with multiple annotations per image.
For this reason the inclusion of classical benchmarks like \cite{cifar,stl-10} is difficult because no multiple annotations or labels are given.
For the CIFAR-10 dataset \cite{cifar}, multiple annotations are only available for the test set as the above mentioned CIFAR-10H dataset \cite{cifar10h} and thus we can evaluate on this subset.
For the STL-10 dataset \cite{stl-10}, only one annotation / label per image is given.
We still include some results of this dataset to illustrate the performance on previous benchmarks.

\subsection{Metrics}
\label{subsec:metric}

We want to measure the quality of classification and clusters over the \certain and \fuzzy data respectively which we assume are better proposals in the annotation process or evaluation by experts.
Based on this reasoning, we decided to use the weighted F1-Score on \certain data and the mean inner distance on \fuzzy data,
The ambiguity is determined by the network output $p_a$.
We define the metrics in detail below and give in \autoref{subsec:consistency} a proof-of-concept for the higher consistency of labels based on proposals selected by the defined metrics.
Common metrics like accuracy are not used because often we have a class imbalance which would lead to misleading results.

During training we do not enforce a balance between \fuzzy and \certain  predictions to keep the required prior knowledge minimal.
This can lead to uninformative metrics and therefore we call a training \emph{degenerated} if no more than 10\% of the validation data are either predicted as \fuzzy or \certain.
We use the \emph{weighted F1-Score} on \certain  images.
We use the weighted version based on the number of images per class to avoid instability due to classes with no or very few \certain  (predicted) images.
For the \fuzzy images, we use the mean inner euclidean distance (\dis) to the centroid on the soft / \fuzzy gt labels.
The metric \disx is based on the soft gt and thus also minimal for classifications of the majority class which allows an evaluation also on classified data.
The equation for a set of clusters of images $X$ is given in \autoref{eq:dis} with sets $C \in X$ as clusters and the corresponding approximated soft label distribution $\hat l_x$ for each image $x \in C$.
The centroid per cluster is given as $\mu_C$.
	\begin{equation}
	\label{eq:dis}
		\begin{split}
	d(X) &:= \frac{1}{|X|} \sum_{C \in X} \frac{1}{|C|} \sum_{x \in C} || \hat l_x - \mu_C ||_2 \text{ with } \\
	\mu_C &:= \frac{1}{|C|} \sum_{x \in C} \hat l_x
	\end{split}
	\end{equation}

We use the vanilla (unchanged) SSL algorithms as baseline experiments.
For these experiments and some ablation experiments, we have no ambiguity prediction $p_a(x)$.
In these cases, we assume all images to be \certain  and use $p_n(x)$ as output.
We often noticed that the classification improved while the clustering degenerated and the other way round.
Therefore, we determine the best performance by looking at the difference \diff\xspace between distance and F1-Score (smaller is better).
It is important to note, that this balancing is arbitrary but we give a proof of concept that the proposals calculated by these metrics lead to more consistent annotations which justifies their definition.
In general, we have 3 runs per setup but we exclude results that degenerate as described above.
We report the best of these runs based on the \diff-score over all non-degenerated runs.
All scores are calculated on the validation data which is in general about 20\% of all the data (see details in \autoref{tbl:datasets}).

\subsection{Implementation Details}
\label{subsec:details}

All methods use the same code base and share major hyperparameters which is crucial for valid comparisons \cite{revisiting-self}.
We use the prior ambiguity $p_A = 0.6$ and loss weights $\lambda_{CE^{-1}} = 10$, $\lambda_f = 0.1$ and $\lambda_s = 0.1$ across all experiments.
It is important to note that we do not use the actual prior probability of ambiguous images $\hat{p_A }$ as given in \autoref{tbl:datasets} because the probability is unknown or would require multiple annotations per image.
We use a constant approximation across all datasets and show in \autoref{subsec:ablation} that this approximation is comparable or even better than $\hat{p_A }$.
This parameter is essential for balancing the certain and \fuzzy images.
The batch size was 64 for all datasets except for the mice bone dataset with a batch size of 8.
The additional losses $L_A$ and $L_S$ are only applied on the unlabeled data while $L_{CE^{-1}}$ is also calculated on the labeled data.
These hyperparameters were determined heuristically on the Plankton Dataset with Mean-Teacher and show strong results across different methods and datasets as shown in \autoref{subsec:results}.
Most likely these parameters are not optimal for an individual combination of a method and a dataset but they show the general usability across  methods and datasets.
We want to show that DC3 can be applied successfully to other datasets without hyperparameter optimization and thus did not investigate all combinations in detail.
Nevertheless, we refer to the supplementary for more detailed insights about individual hyperparameters and the complete pseudo code for the loss calculation.

\subsection{Evaluation}
\label{subsec:results}

The comparison between different SSL algorithms and their extension with DC3 is given in \autoref{tbl:orthogonal}.
The best results were selected as described in \autoref{subsec:metric}.
The complete results and additional plots are given in the supplementary.
We see that DC3 improves the classification and clustering performance across the majority of classes and methods by 5 to 10\%.
\diffx is improved by up to 40\% for 16 out of 19 method-dataset-combinations.
On average, we achieve a 7.6\% higher F1-Score for \certain  classifications  and a 7.9\% lower inner distance for clusterings of \fuzzy images if we look at all non excluded method-dataset-combinations. %
Even on STL-10 (without the possibility to evaluate \fuzzy labels) DC3 creates up to 9\% better classifications.
Overall, we see the most benefit on the Mice Bone and Turkey dataset which we attribute to the worse initial approximation of $\hat l$.
The different vanilla algorithms achieve quite similar results for each dataset.
Only FixMatch achieves a more than 5\% better F1-Score on the curated STL-10 and CIFAR-10H dataset.
In general, we see that DC3 can be beneficially applied to a variety of datasets and methods and  predicts better classifications and more compact clusters.

\tblOrthogonalvTwo

\tblConsistency

\subsection{Proof-of-concept improved data quality}
\label{subsec:consistency}
We show above that DC3 can lead to better classifications and clusters than SSL alone.
In accordance with previous literature \cite{NotBlackAndWhite,foc}, we give a proof-of-concept  in \autoref{tbl:cons} that the annotation process can be improved with cluster-based proposals.
As an SSL algorithm we used Mean-Teacher and for the datasets Plankton, Turkey and CIFAR-10H we used a random subsample of 10\% for the evaluation.
We conducted experiments with a pool of 6 annotators which consisted of domain experts and inexperienced hired workers which were paid a fixed wage per hour.
We assigned 3 annotators from the pool per dataset.
This means that annotator named e.g. A1 might be a different person between datasets in \autoref{tbl:cons}.
We compare the annotations over time from each annotator.
We investigated three different used proposals for the annotation.
The baseline is not using any proposals, the second is using the SSL predictions (classification) and the third is using the DC3 predictions (classification + clusters).
For each cluster, a rough description was given as guidance during the annotation.
After a training phase for the inexperienced annotators, we averaged across three repetitions for every annotator, proposal and dataset combination.

We see a general trend that the consistency improves and the annotation time decreases when proposals are used instead of None.
Using DC3 proposals instead of SSL proposals, either leads to a similar or better consistency while the annotation time is often increased by one or two minutes.
For this improvement, we credit the cleaner and more fine-grained outputs of the network.
The additional verifications of the clusters could lead to the slightly increased annotation time.
The individual benefits vary between the datasets and annotators. %
For example, the gains on the curated CIFAR-10H dataset are lower than on the uncurated Mice Bone dataset.
On average across all annotators and datasets, we achieve an improved consistency of 6.74\%, a relative speed-up of 2.4 and a maximum speed-up of 4.5 with DC3 proposals in comparison to the baseline.

\section{Discussion}

We will discuss the benefits and limitations of our method.
Additional results about the impact of \fuzzy data, the unlabeled data ratio and the interpretabiliy can be found in the supplementary.

\paragraph{Ablation Study}
\label{subsec:ablation}
\tblAblationAverage
We pooled the runs between all methods to evaluate the impact of the individual components of our method DC3 and show the results in \autoref{tbl:ablation-average}.
The method FixMatch and the Mice Bone dataset are excluded from this ablation due to the up to 12 times higher required GPU hours and degenerated runs as before.
Across the datasets, we see the best \diff-scores are achieved by DC3.
The impact of the components varies between the datasets.
We see that CE$^{-1}$ positively impacts the clustering results which confirms the benefit of using CE$^{-1}$ for overclustering \cite{foc}.
CC often reaches a better F1-Score than the baseline and even surpasses DC3 sometimes.
However, the inner distance (\dis) may increase as well.
We conclude that CC and CE$^{-1}$ on their own can lead to improvements but only the combination of both parts results in a stable algorithm across datasets and methods.
Additionally, we see that the number of not degenerated runs is highest with the combination of CE$^{-1}$ and CC.
If we use an realistic amount of ambiguity $\hat p_A$ in each dataset as $p_A$, we see that in general the F1-Score decreases and \dis-score improves.
We attribute this difference to the lower prior ambiguity $p_A$ because DC3 tries to predict more \certain  than \fuzzy images.
This leads to a lower inner distance but also includes more difficult images in the classification of the \certain data.
We believe this parameter is essential for balancing the improvements in the F1- and \dis-score for a specified usecase.
We chose a $p_A$ of 0.6 because we wanted to weight certain and \fuzzy images almost equally but ensure very certain /fewer classifications.

\begin{figure}[htb]
		\begin{subfigure}[c]{0.32\linewidth}
			\includegraphics[width=\textwidth]{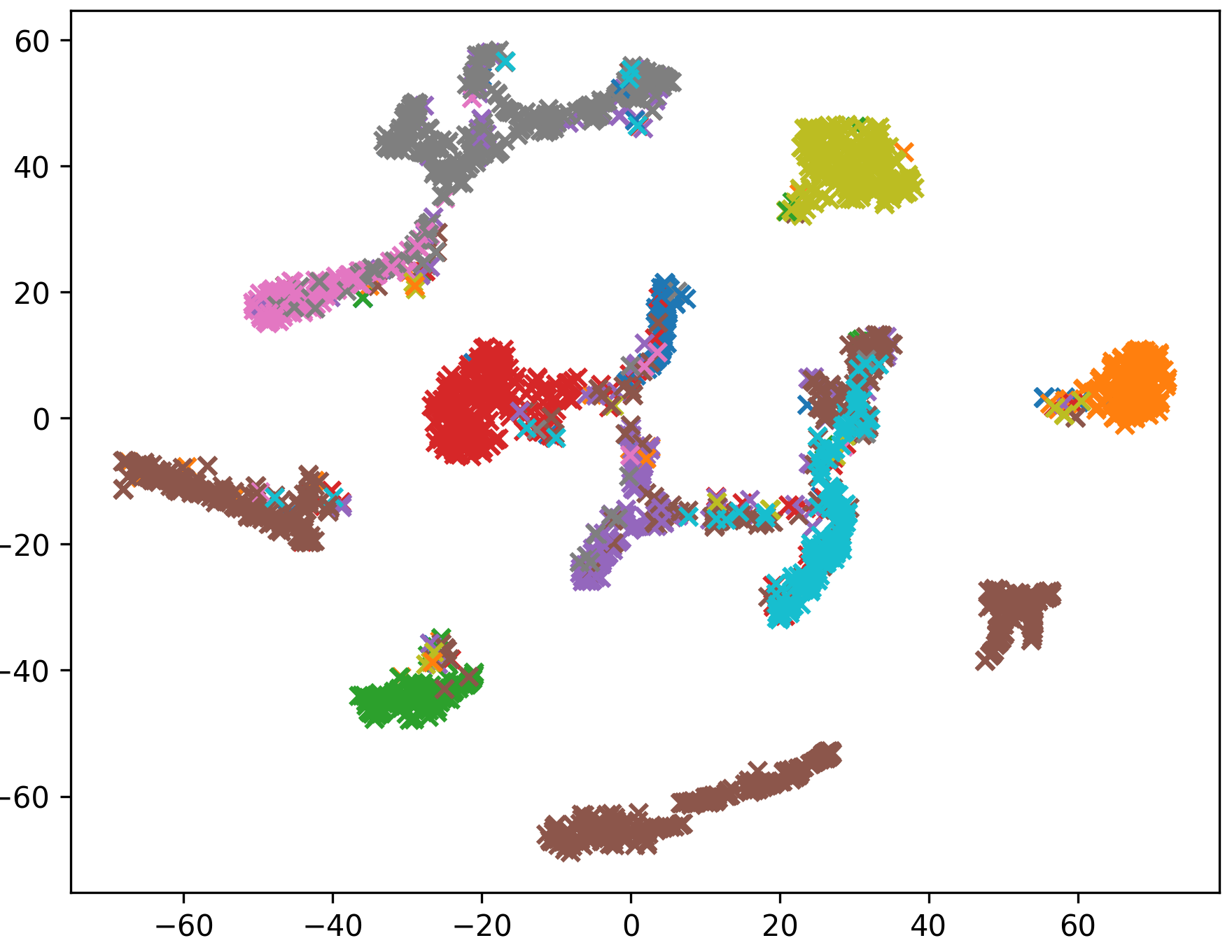}
				\subcaption{GT Classes}
		\end{subfigure}
		\begin{subfigure}[c]{0.32\linewidth}
			\includegraphics[width=\textwidth]{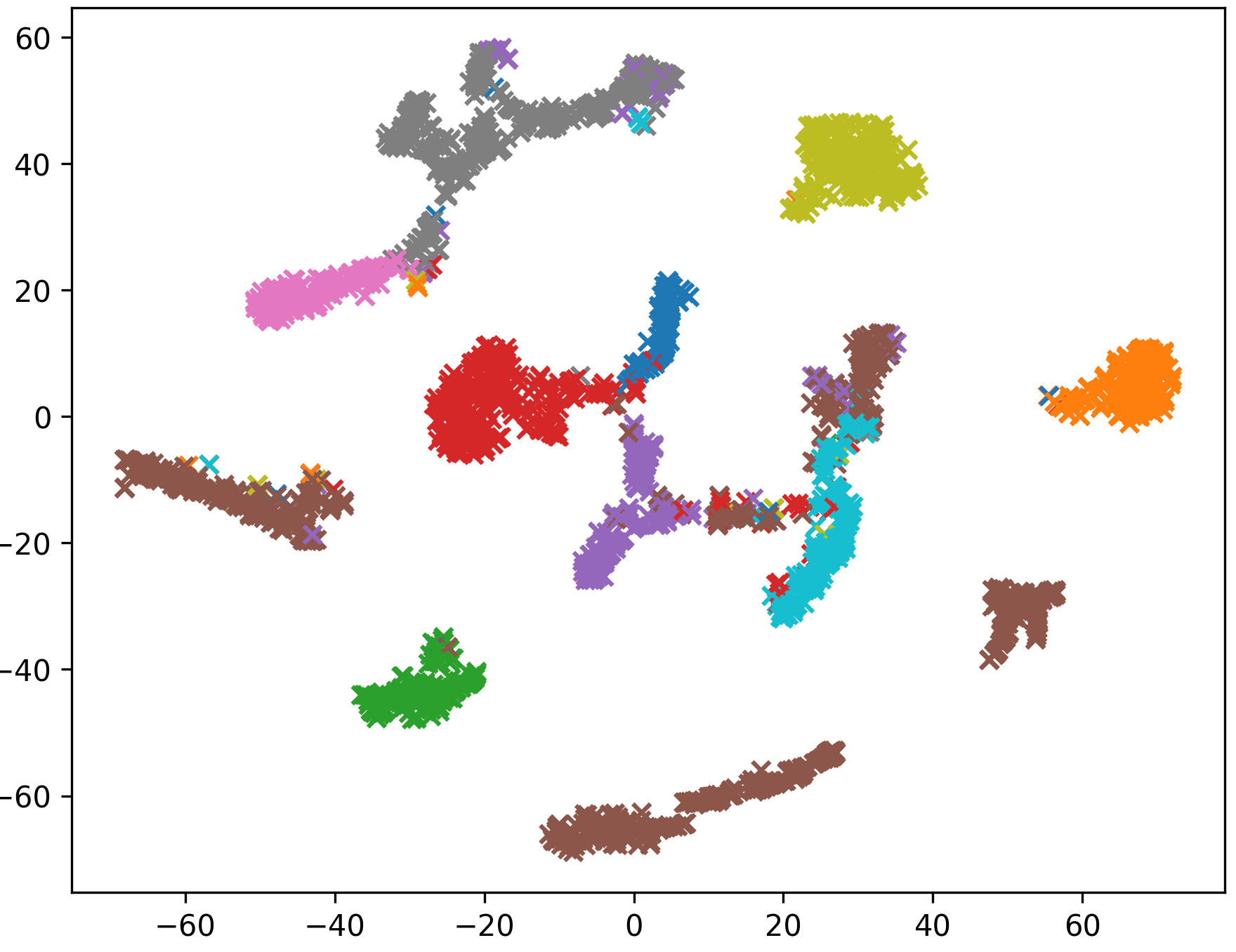}
				\subcaption{DC3 Classes}
		\end{subfigure}
		\begin{subfigure}[c]{0.32\linewidth}
			\includegraphics[width=\textwidth]{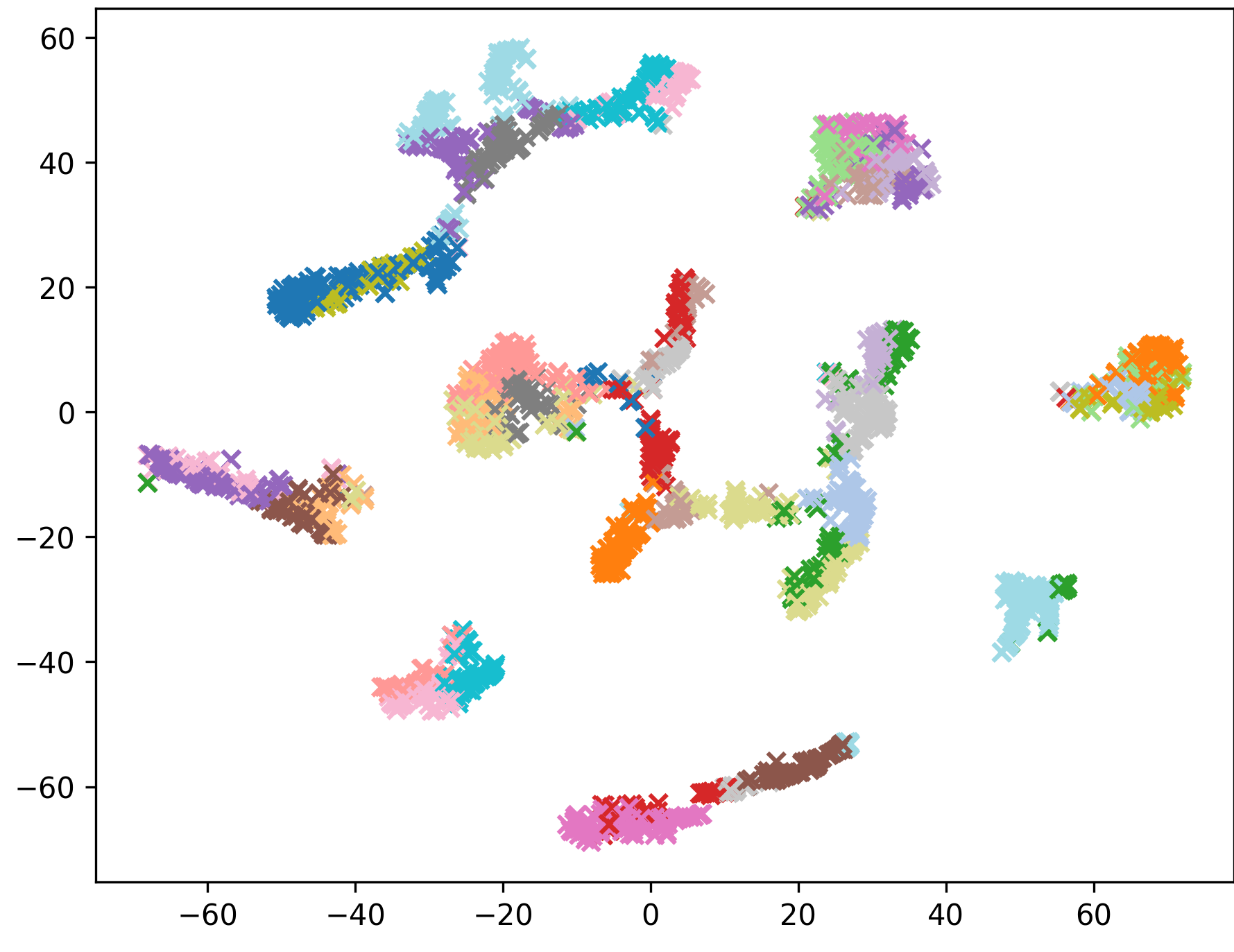}
				\subcaption{DC3 Clusters}
		\end{subfigure}
		\begin{subfigure}[c]{0.32\linewidth}
			\includegraphics[width=\textwidth]{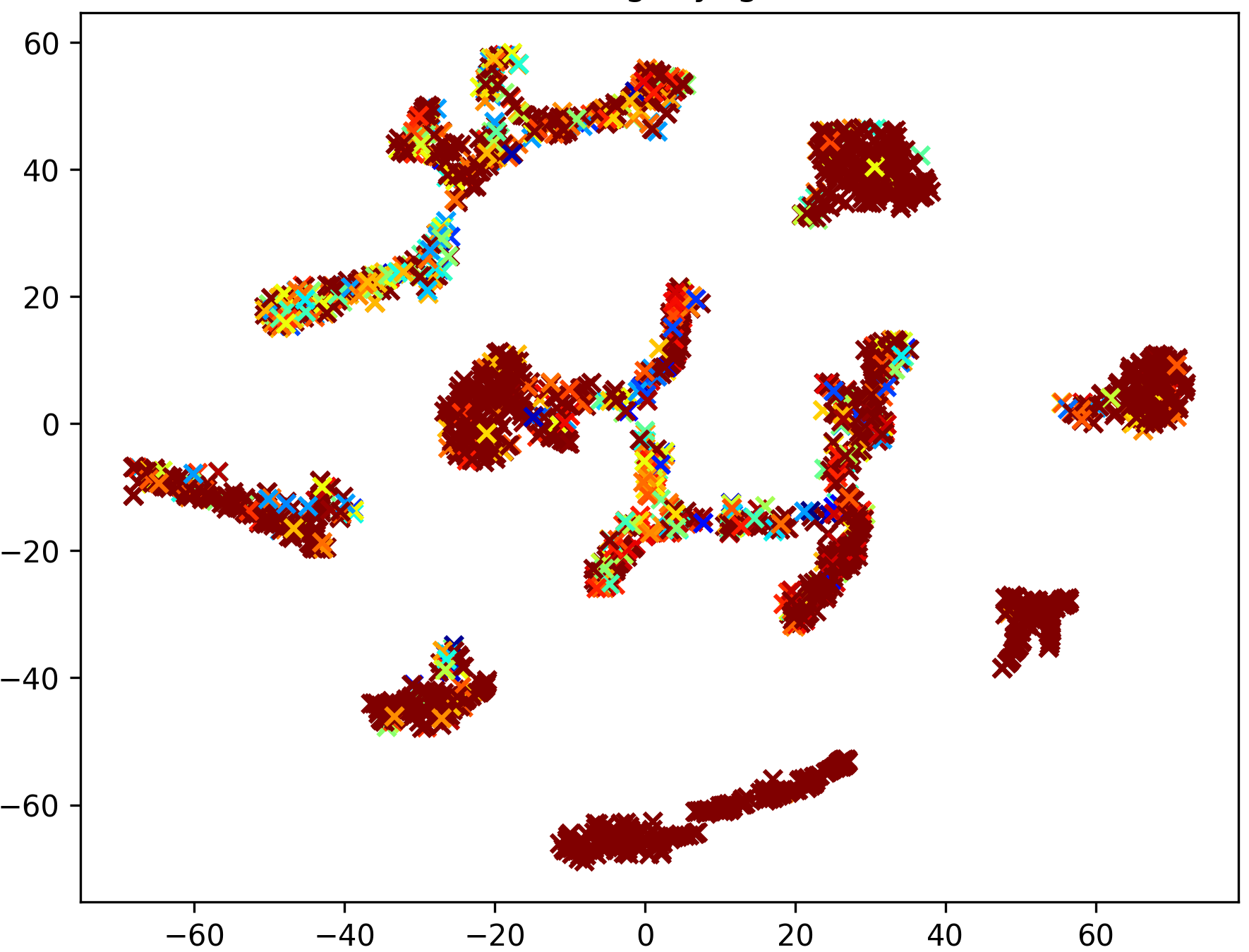}
				\subcaption{GT Ambiguity}
		\end{subfigure}
		\begin{subfigure}[c]{0.32\linewidth}
			\includegraphics[width=\textwidth]{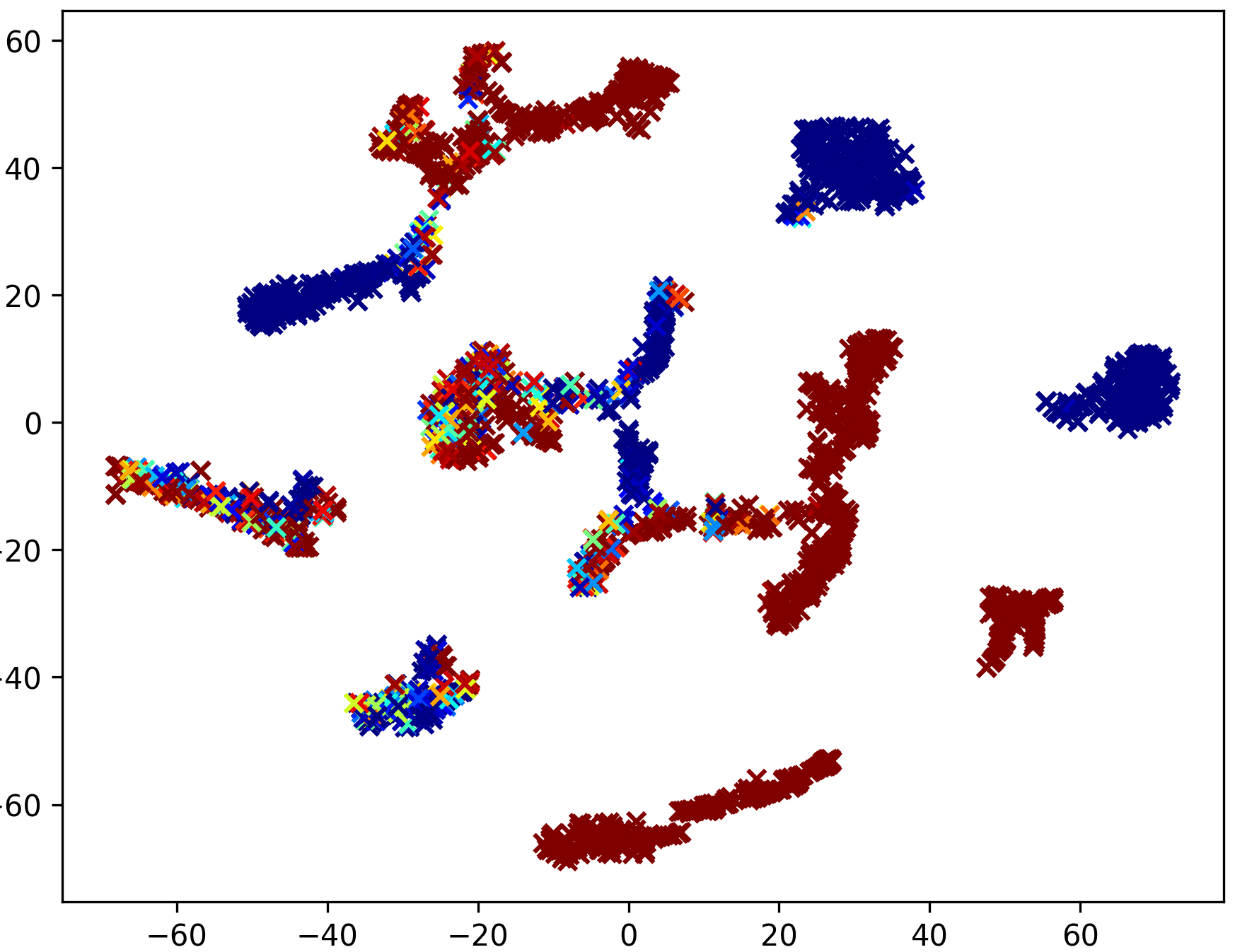}
				\subcaption{DC3 Ambiguity}
		\end{subfigure}
		\begin{subfigure}[c]{0.32\linewidth}
			\includegraphics[width=\textwidth]{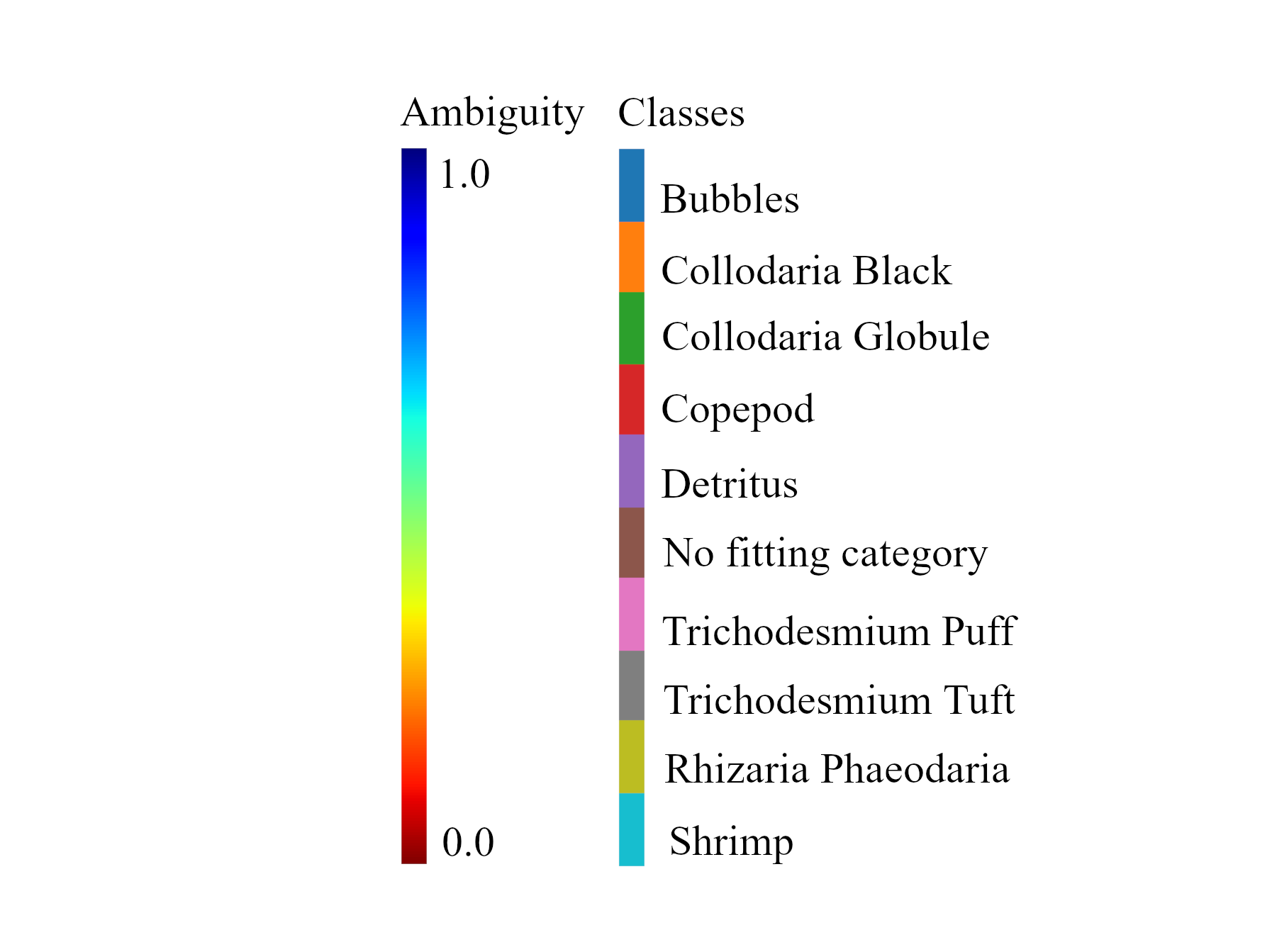}
				\subcaption{Legend}
		\end{subfigure}
		\caption{t-SNE plots for Plankton dataset with Mean-Teacher -- The same color was used 2–3 times for different clusters to ensure distinct colors.}
		\label{fig:tsne}
	\end{figure}

\paragraph{Qualative Analysis with t-SNE}
We qualitatively investigated the t-SNE\cite{van2008tsne} visualizations for one dataset in \autoref{fig:tsne}.
If we compare the predicted (DC3) classes and ambiguity with the ground truth (GT) we see more wrong classifications on ambiguous images and a good estimation of the ambiguity.
DC3 outputs higher ambiguity than expected due to the higher value of $p_A$.
The clusters in (c) partition the feature space in smaller regions.
Overall, we see a better representation of the \fuzzy feature space.

\paragraph{Limitations}
We showed that DC3 generalizes to different SSL algorithms and datasets without hyperparameter changes.
However, the datasets only consists of up to several thousands images.
Due to the required multiple annotations per image for the evaluation it difficult to obtain datasets with millions of images.
In the future, we will investigate the performance on even more and larger datasets to verify the generalizability of DC3.
We focused on improving the classification and clustering and gave a proof-of-concept for the increased consistency of relabeled data.
Due to the required human labor during the relabeling step, we could not investigate the consistency across more datasets and algorithms or investigate the usage of the improved data.
We proposed to improve the annotation process based on human-validated network predictions.
This could introduce a not-desired bias into the data.
This might lead to a negative impact for humans or a group of humans for certain use-cases but
we believe a small bias can be accepted in most applications because it is human controlled and systematically.

\section{Conclusion}

In real-world datasets, we often encounter \fuzzy labels for example due to intra- or interobserver variability.
We propose our method DC3 which is an extension to many SSL algorithms and allows to classify images with \certain  labels and cluster \fuzzy ones.
DC3 also automatically determines which image to treat as \certain  or \fuzzy only based on a given prior probability $p_A$.
On average, we achieve an increased F1-Score of 7.6\% and a lower inner distance of clusters of 7.9\% over all method-dataset-combinations.
We give a proof-of-concept that these improved predictions can be used beneficially as proposals to create more consistent annotations.
On average, we achieve an improved consistency of 6.74\% and a relative speed-up of 2.4 when using DC3 proposals instead of no proposals.
We show that the impact of the different components of DC3, \fuzzy labels can negatively impact the classification performance and the ambiguity prediction can give more insight in the model reasoning.
Therefore, SSL algorithms with DC3 are better suited to handle real-world datasets including \fuzzy labeled images either by an improved classification / clustering or as a proposal during the annotation process with more insight.

\paragraph{Acknowledgements}

	We acknowledge funding of L. Schmarje by the ARTEMIS project (Grant number 01EC1908E) funded by the Federal Ministry of Education and Research (BMBF, Germany).
	R. Kiko also acknowledges support via a “Make Our Planet Great Again” grant of the French National Research Agency within the “Programme d’Investissements d’Avenir”; reference “ANR-19-MPGA-0012”.
	Funds to conduct the PlanktonID project were granted to R. Kiko and R. Koch (CP1733) by the Cluster of Excellence 80 “Future Ocean” within the framework of the Excellence Initiative by the Deutsche Forschungsgemeinschaft (DFG) on behalf of the German federal and state governments.
	Turkey data set was collected as part of the project “RedAlert – detection of pecking injuries in turkeys using neural networks” which was supported by the “Animal Welfare Innovation Award” of the “Initiative Tierwohl”.

{

    \clearpage
    \bibliographystyle{splncs04}
    \bibliography{lib}
}

\appendix

\newpage
{
\centering
\Large
\textbf{A data-centric approach for improving ambiguous labels with combined semi-supervised classification and clustering} \\
\vspace{0.5em}Supplementary Material \\
\vspace{1.0em}
}

\appendix
\setcounter{page}{1}

\section{Further insights into hyperparameters}

During method development, we looked at a larger variety of hyperparameters  for Mean-Teacher \cite{mean-teacher} and the Plankton dataset \cite{foc}.
We found in general that the model was quite robust to changes of individual parameters.
The most impact we noticed from design decision where and where not a gradient is propagated.
For example, if we would propagate the error along the pseudo labels for the ambiguity loss calculation the system degenerates almost always.
Aside from these design decisions, the weight for $\lambda_{CE^{-1}}$ had the most impact.
While we use the same value for the labeled and the unlabeled data, we see slight evidence that a lower value could be beneficial on the labeled data.
Moreover, slightly lower or higher weights for the other hyperparameters showed promising results under certain circumstances.
As stated in the paper, we aimed at providing general parameters across different datasets and methods and thus did not fine-tune the parameters to a specific combination.

\section{Pseudocode}

In our code block below, we give the main parts of our proposed method as pseudocode.
The code is similar to python and Tensorflow code.
In the following, we will describe the used parameters and methods.
\verb|tf| is an abbrevation for tensorflow and refers to that function.
\verb|prob_ambiguous| is the output $p_a(x)$ for a complete batch.
\verb|logits_x_over|, \verb|logits_u_over| and \verb|logits_u2_over| are the overclustering outputs $p_o(x)$ for a complete batch for the labeled data, the unlabeled data and the possible additional second unlabeled input respectively.
The labels for the labeled data are given in \verb|l|.
\verb|logits_u| is the output $p_n(x)$ for a complete batch of unlabeled data.
The parameters \verb|prior_ambiguity|, \verb|wou|,\verb|wa| and \verb|ws| correspond to the hyperparameters $p_A, \lambda_{CE^{-1}}, \lambda_a $ and $\lambda_s$ in the paper respectively.
The parameter \verb|wol| is the weight $\lambda_{CE^{-1}}$ on the labeled data.
The parameter \verb|loss_tensor| is SSL loss as tensor ( $L_{SSL}$).
The function \verb|threshold()| thresholds the elements of the given vector (first argument) based on the given threshold (second argument).
The functions \verb|inverse_ce()|, \verb|ce()| and  \verb|be()| calculate the loss value for inverse cross-entropy (CE$^{-1}$), cross-entropy and binary cross-entropy respectively.
The function \verb|get_different_logits()| selects from the given batch logits (first argument) one logit for each image in the batch.
A logit is randomly selected of all logits in the batch which to do not share the same label (\verb|l|) or the same pseudo label based on \verb|logits_u|.
The function \verb|get_pseudo_ambiguity_labels()| gets pseudo labels for every ambiguity prediction $p_a(x)$ based on the given prior ambiguity estimate $p_A$ as described in the main paper.

The different outputs can easily be realized in a model by extending the dense output layer to the sum of the number of classes and the  number of output clusters.
Before calculating the loss or applying softmax activation the output can be separated in the desired input to our method.

\begin{lstlisting}
def calculate_loss(prob_ambiguous, l,
    logits_x_over, logits_u, logits_u_over, logits_u2_over,
    prior_ambiguity, wou, wol, wa, ws,
    loss_tensor):

    # stop gradient on ambiguity scale
    ambiguous_scale = tf.stop_gradient(prob_ambiguous)
    certain_scale = 1 - ambiguous_scale


    pseudo_labels = tf.stop_gradient(tf.nn.softmax(logits_u))
    args_pseudo = tf.argmax(pseudo_labels, axis=1)
    pseudo_mask = threshold(pseudo_labels, 0.95)

    #  get different image based on label or pseudo-label
    # for elements in the batch from the batch
    logits_x_over_inverse = get_different_logits(logits_x_over, l)
    logits_u_over_inverse = get_different_logits(logits_u_over, l)

    loss_xeil = inverse_ce(tf.nn.softmax(logits_x_over),
        tf.nn.softmax(logits_x_over_inverse))
    loss_xeil = tf.reduce_mean(loss_xeil)

    loss_xeiu = inverse_ce(tf.nn.softmax(logits_u_over),
        tf.nn.softmax(logits_u_over_inverse))
    loss_xeiu = tf.reduce_mean(loss_xeiu * pseudo_mask * certain_scale)

     # use pseudo labels based on the number of ambiguous elements
     # in each batch to calculate ambiguity loss
    pseudo_ambiguity_label =
        get_pseudo_ambiguity_labels(prob_ambiguous, prior_ambiguity)
    loss_ambiguity = be(tf.stop_gradient(pseudo_ambiguity_label),
        prob_ambiguous)
    loss_ambiguity = tf.reduce_mean(loss_ambiguity)

    # calculate similarity loss
    # use ce to add entropy based on the logits u over
    sim_loss = ce(tf.nn.softmax(logits_u_over),
        tf.nn.softmax(logits_u2_over))
    sim_loss = tf.reduce_mean(sim_loss * ambiguous_scale)

    loss = loss_tensor * certain_scale + wou * loss_xeiu + wol * loss_xeil
        + wa * loss_ambiguity + ws * loss_ambiguous_similarity

    return loss
\end{lstlisting}

\section{Additional Results}

\paragraph{Impact of ambiguous labels}
\label{subsec:ambiguous}
We stated that high quality labels lead to better model training \cite{are_we_done} and verify this statement on the Plankton and CIFAR-10H datasets in \autoref{tbl:fuzziness}.
We see for all supervised and semi-supervised methods that used training labels based on the complete distribution of $\hat l$ (argmax $\hat l$, column A) leads to an improvement of up to 10\% in comparison to sampling the training label from $\hat l$ (column A1).
We used the approximation based on a sample from $\hat l$ because we normally would have access to $\hat l$ only at a high cost.
If we remove the \fuzzy images entirely from the dataset (column C), the results improve again by 8 to 15\%.
This indicates that \fuzzy images are a major issue during the training process.

\tblImpactFuzziness

\picTwoExtended{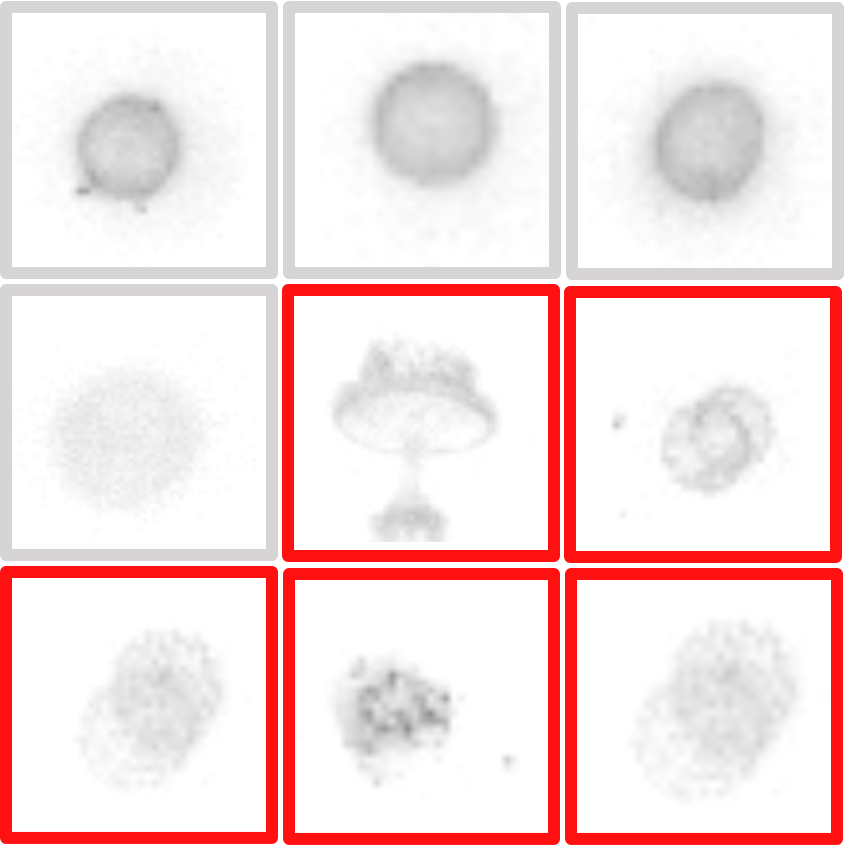}{Plankton - Collodaria globule}{CIFAR-10H - Deer}{Mice Data - Dissimilar}{ Turkey - Not Injured}{Qualitative Results for selected classes across different confidences and ambiguity predictions --
Wrong classifications based on the normal head are highlighted in red. The ground-truth class is given in the subcaption.}

\paragraph{Interpretability}
\label{subsec:interpret}
Many SSL algorithms interpret the probability of the largest value of $p_n(x)$ as confidence \cite{guo2017ece}.
We qualitatively illustrate in \autoref{fig:fuzziness-extended} that using our ambiguity prediction $p_a(x)$ can lead to better interpretability and fewer errors.
 We show 6 randomly picked examples for selected classes across the datasets and extended results in the supplementary.
 The images in each row have a similar value for $p_n(x)$ and $p_a(x)$.
 The first row presents highly confident predictions on certain predicted images and shows no errors in the given random picks.
 The middle row shows highly confident predictions on \fuzzy predicted images.
 Some of these images are false and would lower the performance without the additional ambiguity prediction.
 The last row shows non-confident or uncertain ($0.4 < p_a(x) < 0.6$) predictions which are often wrong.

\paragraph{Impact of unlabeled data ratio}

In this study, we fixed the supervision to a fixed ratio to simplify the analysis.
However, we conducted some preliminary studies on a variant of the MiceBone dataset and show the results in \autoref{tbl:ablation}.
Be aware that the numbers are not directly comparable due to a different data split algorithm.
We see that DC3 improves the results over all supervision percentages.
It seems also be more robust to less data than SSL alone, but additional data is required to confirm this trend.

\begin{table}[tb]

	\caption{
	    Impact of unlabeled data ratio --
	    The first column unlabeled data ratio.
	     Better results compared to baseline are bold.
	}
	\label{tbl:ablation}

	\resizebox{\linewidth}{!}{
			\begin{tabular}{l l c c c c c c c c c}
				\toprule
			& & \multicolumn{2}{c}{F1}  & \multicolumn{2}{c}{$d$}  & \multicolumn{2}{c}{$d-$F1} & \\

\cmidrule(r){3-4} \cmidrule(r){5-6} \cmidrule(r){7-8}

				& & best & mean $\pm$ std  &  best & mean $\pm$ std   &  best & mean $\pm$ std &  \\

			\midrule

				\rowcolor{LightGray}

			\textbf{10\%} & 	Mean-Teacher  &
				0.5850& 0.5627 +- 0.0181 & 0.4639& 0.4889 +- 0.0275 & -0.1095& -0.0738 +- 0.0304\\

				& + DC3   &
				\textbf{0.6044}& 0.5277 +- 0.0655 & \textbf{0.2590} & \textbf{0.3011 +- 0.0319} &\textbf{ -0.3455} & \textbf{-0.2266 +- 0.0879}  \\

			\rowcolor{LightGray}
			\textbf{20\%} &	Mean-Teacher  &
				 0.5643& 0.5611 +- 0.0044 &0.4721& 0.5157 +- 0.0326 & -0.0827& -0.0454 +- 0.0284\\

			&	 + DC3   &
				\textbf{0.6111} & 0.5380 +- 0.0519 & \textbf{0.2175} & \textbf{0.3772 +- 0.1229} &  \textbf{-0.3936} & \textbf{-0.1608 +- 0.1731} \\
			\rowcolor{LightGray}
			\textbf{50\%} &	Mean-Teacher  &
				0.7348& 0.6700 +- 0.0856 & 0.4388& 0.4603 +- 0.0270 & -0.2910& -0.2097 +- 0.1124
\\

			&	 + DC3   &
				0.6862& \textbf{0.6839 +- 0.0023} & \textbf{0.3298} & 0.4601 +- 0.1303 & \textbf{-0.3518} & \textbf{-0.2238 +- 0.1279}  \\
			\rowcolor{LightGray}
			\textbf{100\%} &	Mean-Teacher  &
				0.7254& 0.6524 +- 0.0645 & 0.3875& 0.4373 +- 0.0355 & -0.3379& -0.2150 +- 0.0971\\

			&	 + DC3   &
				0.5692 & 0.6017 +- 0.0325 & \textbf{0.1147} & \textbf{0.2995 +- 0.1848} & \textbf{-0.4545} & \textbf{-0.3022 +- 0.1523} \\

				\midrule

		\end{tabular}
	}

\end{table}

 \paragraph{Social Impact Discussion}
Improving the annotation process by increasing the data quality and reducing the required time leads indirectly to improved performance in broad range of deep learning applications.
However, these improvements are achieved by leveraging the confirmation bias for given proposals.
We propose that network predictions are validated by a human but introducing a bias without considering the consequences could lead to undesirable behaviour in concrete cases.
For example, if people are classified / graded based on provided proposals a negative bias could be introduced for certain people.
In such a case, the user should consider investigating more annotation resources into an unbiased consensus process.
We believe a small bias can be accepted in most applications because it is human controlled and systematically.

\section{Extended results}

In this subsection, we give the complete result tables used for the tables in the main paper.
The definitions of the metrics are given in the main paper.
Detailed scatter plots are given in \autoref{fig:scatter}.
The individual tables are \autoref{tbl:ablation-ce}, \autoref{tbl:ablation-mean}, \autoref{tbl:ablation-pi}, \autoref{tbl:ablation-pseudo} and \autoref{tbl:ablation-fixmatch}.

\picTwo{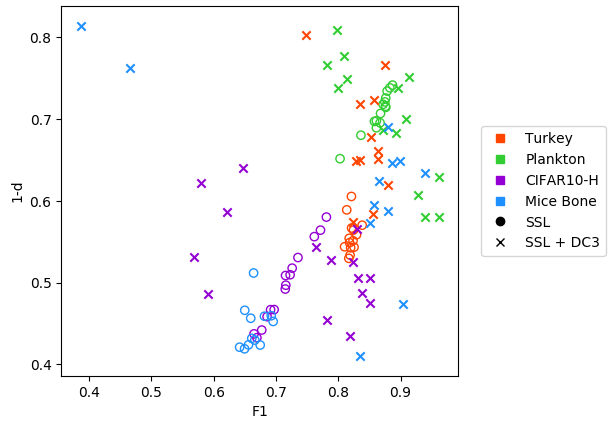}{SSL vs. SSL + DC3}{Ablation study}{Each datapoint represents an independent training run depending on the weighted F1-Score (F1) and the mean inner distance (\dis). The color and marker types define the used dataset and method respectively.
For the ablation, we evaluate DC3 with the loss $CE^{-1}$ (CE-1), only the classification and clustering without the loss (CC) and the combination (DC3).
We usually use the prior probability for the ambiguity ($p_A$,pA) of 60\% but show also an ablation with the realistic prior $\hat p_A$ from \autoref{tbl:datasets}.
}

\tblAblationCE

\tblAblationMean

\tblAblationPi

\tblAblationPseudo

\tblAblationFixmatch

\end{document}